\documentclass[11pt]{article}

\usepackage[final]{acl}

\usepackage{times}
\usepackage{latexsym}
\usepackage[T1]{fontenc}
\usepackage[utf8]{inputenc}
\usepackage{microtype}
\usepackage{inconsolata}
\usepackage{graphicx}
\usepackage{booktabs}
\usepackage{amsmath}
\usepackage{multirow}

\usepackage{enumitem}
\usepackage{cleveref}
\usepackage{xurl}
\usepackage[table,xcdraw]{xcolor}
\usepackage{longtable}
\usepackage{makecell}
\usepackage{twemojis}
\usepackage{tabularx}
\usepackage{fontawesome}
\usepackage{cuted}

\title{The Language--Energy Divide:\\
Measuring Energy Costs of Multilingual LLM Inference}

\author{
    Naihao Deng$^{\twemoji{peach}}$
    \quad
    Alissa Shen$^{\twemoji{peach}}$ \quad
    Yiming Feng$^{\twemoji{peach}}$ \quad
    {\bf Joan Nwatu$^{\twemoji{peach}}$ \quad}
    {\bf Jae-Won Chung$^{\twemoji{peach}}$ \quad}\\
    {\bf Mosharaf Chowdhury$^{\twemoji{peach}}$ \quad}
    {\bf Yulong Chen$^{\twemoji{cherries}\twemoji{blueberries}}$\quad}
    {\bf Rada Mihalcea$^{\twemoji{peach}}$}
    \\
    $^{\twemoji{peach}}$University of Michigan\quad
    $^{\twemoji{cherries}}$University of Aberdeen\quad
    $^{\twemoji{blueberries}}$University of Cambridge
    \\
    \texttt{\{\href{mailto:dnaihao@umich.edu}{dnaihao}, \href{mailto:mihalcea@umich.edu}{mihalcea}\}@umich.edu}\quad\texttt{\href{mailto:yulongchen1010@gmail.com}{yulong.chen@abdn.ac.uk}}
}

\begin{document}
\maketitle

\begin{strip}
\centering
\small
\begin{tabular}{@{}l l l@{}}
\faGlobe    & \textbf{Project page:} & \url{https://lit.eecs.umich.edu/language-energy-divide/} \\
\faGithub   & \textbf{Code:}         & \url{https://github.com/MichiganNLP/language-energy-divide} \\
\faDatabase & \textbf{Data:}         & \url{https://huggingface.co/datasets/MichiganNLP/language-energy-divide} \\
\end{tabular}
\end{strip}

\begin{abstract}
Large language models (LLMs) are increasingly deployed in multilingual settings, yet the energy costs of serving these models across different languages remain poorly understood.
We present a systematic study of inference energy consumption across languages with ML.Energy framework~\citep{chung2026the}.
We find striking disparities: \textit{energy consumption per output token varies by up to 8.3$\times$ across languages, while total energy for a fixed set of requests varies by up to 179$\times$ between the cheapest (English, 17.6~kJ) and the most expensive (Pashto, 3{,}147~kJ) languages}.
Our analysis shows that this disparity is driven by two compounding factors: (1) higher \textit{per-token energy costs} for languages using complex or rare scripts, and (2) more tokens generated for low-resource languages.
Moreover, we find a double cost + performance penalty: languages with the highest energy footprints also tend to achieve the lowest task accuracy.
We reveal that the energy divide persists across models, hardware, and tasks, suggesting a systemic energy inequity in multilingual LLM deployment.
Finally, we recommend that the community treat energy as a first-class evaluation axis, extend reporting checklists and model cards to include it, and adopt deployment-side mitigations for better energy efficiency.
\end{abstract}

\section{Introduction}

\begin{figure*}[t]
  \centering
  \includegraphics[width=0.8\textwidth]{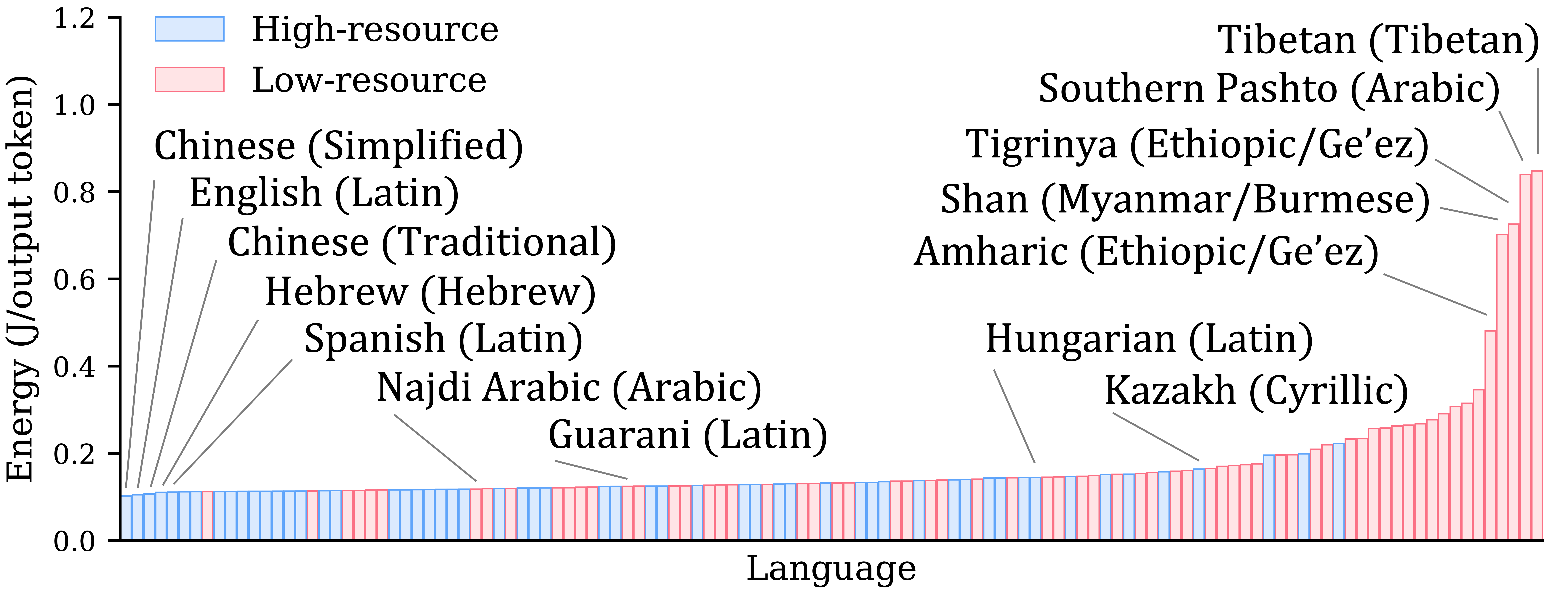}
  \caption{\textit{Energy per output token} across 122 languages.
  Languages are sorted by energy cost and colored by resource level.
  The distribution is right-skewed, with the cluster of low-resource languages exceeding $0.4$~J/token on the right.
  }
  \label{fig:energy_per_token}
\end{figure*}

The rapid adoption of large language models (LLMs) has raised urgent questions about their environmental impact and computational costs~\citep{patterson2022carbon, cbre2023,cbre2024,cbre2025,mckinsey2023,mckinsey2024,zuckerberg-ai-energy,electricity-iea25,bloombergnef25}.
As the field gradually shifts from research prototypes to deployed infrastructure, the dominant evaluation paradigm that focuses on task performance may be insufficient in capturing what matters for real-world use.

Multilingual benchmarks~\citep{liang-etal-2020-xglue, bandarkar-etal-2024-belebele, shi2023language, xuan-etal-2025-mmlu} in particular focus almost exclusively on per-language task performance, while overlooking the energy cost.
On the other hand, prior work has shown that low-resource languages require more tokens to express semantically equivalent content \citep{ahia-etal-2023-languages,petrov2023language}, suggesting potentially unequal computational and energy costs.
However, existing analyses primarily rely on token-level proxies such as sequence length or token counts, while the actual hardware-level energy consumption remains underexplored.

\begin{figure*}[t]
  \centering
  \begin{minipage}{.45\linewidth}
    \centering
    \includegraphics[width=\textwidth]{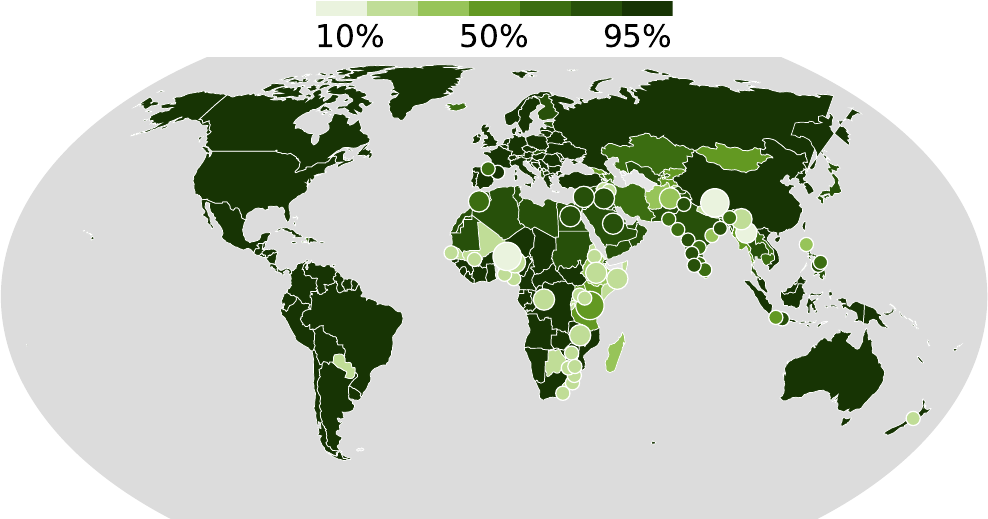}
    \vspace{-0.7cm}
    \caption*{(a) Accuracy.}
  \end{minipage}
  \begin{minipage}{.45\linewidth}
    \centering
    \includegraphics[width=\textwidth]{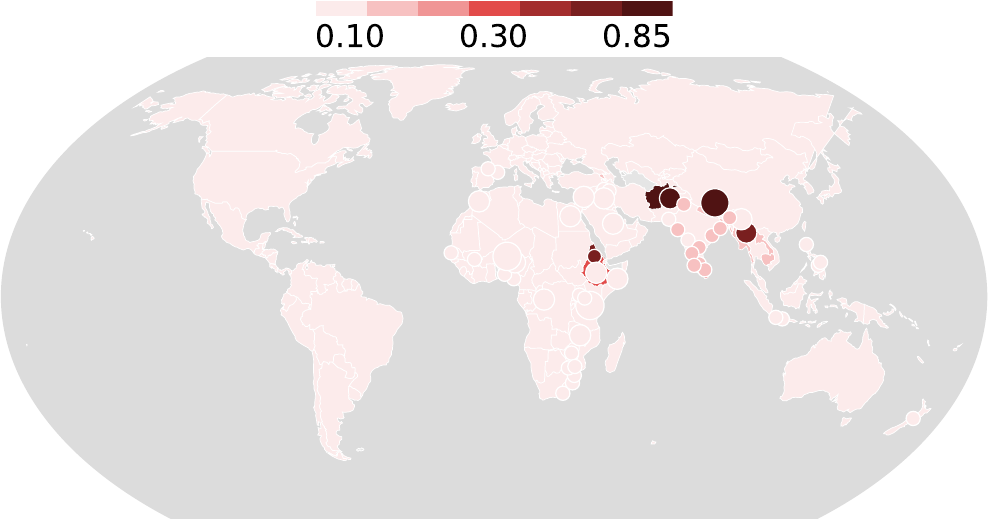}
    \vspace{-0.7cm}
    \caption*{(b) Energy/token (J/tok).}
  \end{minipage}\hfill
  \begin{minipage}{.45\linewidth}
    \centering
    \includegraphics[width=\textwidth]{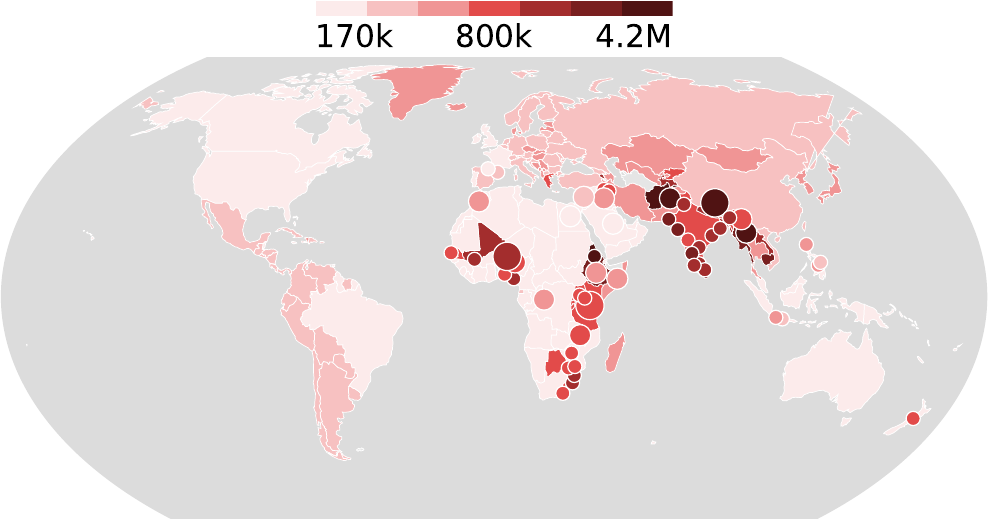}
    \vspace{-0.7cm}
    \caption*{(c) Output token number.}
  \end{minipage}
  \begin{minipage}{.45\linewidth}
    \centering
    \includegraphics[width=\textwidth]{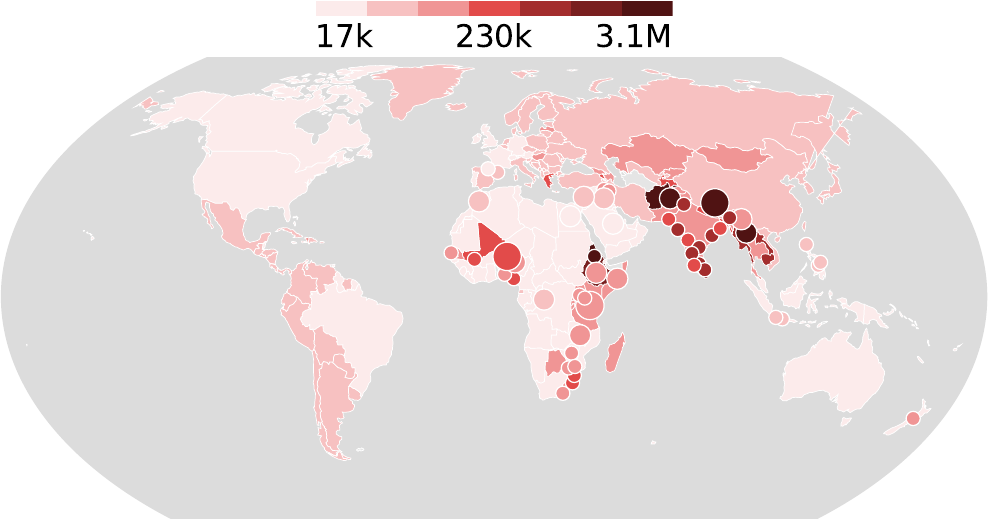}
    \vspace{-0.7cm}
    \caption*{(d) Total energy (J).}
  \end{minipage}
  \vspace{-0.2cm}
  \caption{Geographic distribution of task accuracy and energy consumption (full results in \Cref{tab:full_results}).
  We note that the figure is intended as a qualitative geographic summary of our result.}
  \label{fig:world_maps_all}
\end{figure*}

In this work, we address the following research question: \textit{Does the language of a query systematically affect the energy cost of LLM inference, and if so, by how much?}
If answering a question in a low-resourced language incurs significantly more energy as compared to answering the same question in English, LLM providers can face economic pressure to deprioritize low-resource languages, deepening the digital divide.
To answer this question, we adopt the ML.Energy benchmarking framework~\citep{chung2026the} to measure the energy consumption of several models across all 122 languages of the Belebele reading comprehension dataset~\citep{bandarkar-etal-2024-belebele}, and the 8 languages from the translated GSM8K math reasoning dataset \citep{cobbe2021gsm8k} and the LM-Arena open-ended chat \citep{chiang2024chatbot}.
Within each model and hardware configuration, all languages are evaluated under identical inference conditions.

We experiment across model families, GPUs, batch sizes, and tasks, and uncover several interconnected phenomena.
First, \textit{per-token energy} costs vary by up to 8.3$\times$ across languages, revealing substantial cross-lingual inequity in computational cost.
Second,  low-resource languages typically incur both the highest per-token energy costs and the most inflated token counts, yielding total energy costs up to 179$\times$ the total energy of English. At the same time, we also find that these languages simultaneously achieve low task accuracy, leading to a double cost + performance penalty.
These disparities persist across all configurations we test, indicating a systemic energy inequity in multilingual LLM deployment.
\Cref{fig:world_maps_all} provides a qualitative geographic summary of our results.
We close by arguing that current performance-centered evaluation conceals these disparities, and by laying out concrete recommendations for the community, including treating energy as a first-class evaluation axis, extending checklists and model cards to include energy, and adopting deployment-side mitigations for better energy efficiency.

\section{Related Work}

\paragraph{LLM energy and efficiency.}
A substantial body of work has studied the computational cost of training LLMs~\citep{schwartz2020green, patterson2022carbon}, and more recently, the inference cost at scale~\citep{desislavov2023trends, chung2026the}.
Efforts such as Zeus~\citep{you2023zeus}, CodeCarbon~\citep{codecarbon}, and more recent inference-focused benchmarks~\citep{niu2026tokenpowerbench} provide tools for measuring and tracking energy at the hardware level.
Leveraging these tools, \citet{chung2026the, chung2026where} measure the LLM energy consumption across a wide spectrum of tasks, and \citet{luccioni2024power} compare energy footprints across model families and modalities.
However, prior work has largely focused on efficiency in the context of a single language (typically English)~\citep{chung2026the}, leaving multilingual inference costs unexplored.
In this work, we leverage the existing tools to measure energy consumption in the multilingual setup, systematically measuring energy consumption across 122 languages.

\paragraph{Multilingual NLP and resource disparities.}
The gap in NLP performance between high- and low-resource languages is extensively documented~\citep{conneau-etal-2018-xnli, joshi-etal-2020-state, liang-etal-2020-xglue, hu2020xtreme, shi2023language, chen-etal-2023-revisiting, yan2024gpt, xuan-etal-2025-mmlu, zan2026multiswebench}, and recent multilingual benchmarks have substantially broadened coverage~\citep{bandarkar-etal-2024-belebele, singh-etal-2024-aya, pomerenke2025ai, romanou2025include}.
In this paper, we leverage the multilingual benchmarks constructed by our community~\citep{bandarkar-etal-2024-belebele} for the purpose of energy measurement.
Recent work shows that tokenizers are themselves a source of cross-lingual inequity.
Scripts with high character-to-token ratios pay more per query in API-priced systems~\citep{rust-etal-2021-good, ahia-etal-2023-languages, petrov2023language}, with up to 15$\times$ token inflation relative to English for some morphologically rich or non-Latin-script languages~\citep{petrov2023language}.
While model-quality and tokenizer disparities for low-resource languages are well-known, their compute and energy implications have not been studied at scale.

\section{Preliminaries}\label{sec:preliminaries}

\paragraph{Energy Consumption and Measurement.}
Every invocation of computation and memory movement on hardware consume both time and energy.
Energy consumption is determined not only by \emph{what} is computed, but also \emph{how} it is computed on the hardware, requiring real measurement to derive insights.
Similar to time, modern hardware expose power and/or energy consumption via a suite of hardware counters specific to each vendor, which can be sampled to measure the energy consumption of workloads.
In this work, we use the Zeus library~\citep{you2023zeus,zeus-github} to measure and report the energy consumption of NVIDIA GPUs.
Hardware counters are known to be accurate within a 5\% error margin~\citep{nvml} and have also been verified independently~\cite{nvml-verified}.

\paragraph{Multilingual Dataset.}
We run our tests and analyses on multilingual datasets. Belebele \citep{bandarkar-etal-2024-belebele} is a multi-choice reading comprehension benchmark, where each item pairs a short passage with a four-option question, and a model must select the correct answer.
Its items are parallel across all 122 languages (the same passages and questions, translated).
We use all of the 900 questions per language from Belebele.

For analyses requiring task diversity, we also use GSM8K \citep{cobbe2021gsm8k}, which is a benchmark of grade-school math word problems that require multi-step arithmetic reasoning; and LM-Arena \citep{chiang2024chatbot}, which is a collection of real, open-ended user prompts to chat assistants with no reference answer.
As neither of these latter datasets is natively multilingual, we machine-translate both so that all three tasks span the identical 8-language subset, including four high-resource languages of English (Latin), Chinese (Simplified), Russian (Cyrillic), French (Latin), and four low-resource languages of Southern Pashto (Arabic), Tigrinya (Ethiopic/Ge'ez), Shan (Myanmar/Burmese), Tibetan (Tibetan).
\footnote{We note that names in parentheses denote the writing system (script), not the language or language family. 
Throughout the paper, we identify languages by their FLORES-200 codes~\citep{costa2022no}, which pair an ISO~639-3 language code with an ISO~15924 script code.
For instance, \texttt{pbt\_Arab} is Southern Pashto and \texttt{pes\_Arab} is Western Persian, both written in Arabic-derived scripts. 
We also note that a shared script code does not imply any linguistic grouping.
For instance, Pashto and Persian are Iranian languages, unrelated to Arabic.}
We provide the details of how we process each dataset in Appendix~\ref{appsec: experimental-setups}.

\paragraph{Language Resource Classification.}
We use the No Language Left Behind (NLLB) taxonomy~\citep{costa2022no} to classify the resource level of the studied languages.
Of the 122 Belebele languages, 56 are classified as high-resource and 66 as low-resource.

\section{Results and Analysis}

\begin{table}[t]
  \centering
  \small
\resizebox{\linewidth}{!}{
  \begin{tabular}{clrrrrr}
    \toprule
    \textbf{Res.} & \textbf{Lang.} & \textbf{J/tok} & \textbf{Tok/R} & \textbf{E\textsubscript{tot} (J)} & \textbf{$\times$En} & \textbf{Acc} \\
    \midrule
    \multirow{4}{*}{High} & English\textsubscript{(Latin)} & 0.10 &   186 &     17{,}588 & 1.0$\times$ & 94.6 \\
         & Chinese\textsubscript{(Simplified)} & 0.10 &   403 &     36{,}983 & 2.1$\times$ & 89.6 \\
         & French\textsubscript{(Latin)} & 0.11 &   288 &     29{,}420 & 1.7$\times$ & 91.7 \\
         & Russian\textsubscript{(Cyrillic)} & 0.12 &   411 &     43{,}560 & 2.5$\times$ & 90.9 \\
    \midrule
    \multirow{4}{*}{Low}  & Tigrinya\textsubscript{(Ethiopic/Ge'ez)}  & 0.73 & 3{,}695 & 2{,}413{,}328 & 137$\times$ & 25.3 \\
         & Shan\textsubscript{(Myanmar/Burmese)} & 0.70 & 4{,}667 & 2{,}948{,}722 & 168$\times$ & 10.6 \\
         & Tibetan\textsubscript{(Tibetan)} & 0.85 & 3{,}937 & 3{,}002{,}867 & 171$\times$ & 21.9 \\
         & Southern Pashto\textsubscript{(Arabic)} & 0.84 & 4{,}164 & 3{,}146{,}541 & 179$\times$ & 40.4 \\
    \bottomrule
  \end{tabular}}
  \caption{Energy gap for Qwen3-8B on Belebele.
  ``Res.'', ``Lang.'', ``J/tok'', ``Tok/R'', ``E\textsubscript{tot} (J)'', ``$\times$En'', and ``Acc'' refer to resource level, language, energy per output token, output tokens per response, total energy in J, total energy relative to English, and accuracy, respectively.
  }
  \label{tab:energy_decomposition}
\end{table}

\paragraph{Common Setup.}

Each language is evaluated under identical conditions, served via a vLLM~\citep{kwon2023efficient} inference server on an HPC cluster.
Following the ML.ENERGY Benchmark~\citep{chung2026the}, we identify a \textit{steady-state} period defined as the window during which the server's batch size is saturated at its maximum configured value, which is the period most representative of long-term deployment conditions.
Unless otherwise noted, all per-token energy values reported in this paper are steady-state values; this ensures that languages with short, ``well-behaved'' runs (e.g., English) are not penalized for cold-start overhead relative to languages whose runs are dominated by extended token generation.

\subsection{Energy Varies Substantially Across Languages}
\label{subsec: energy-vary-across-languages}

\paragraph{Setup.}
We run our measurements using Qwen3-8B~\citep{yang2025qwen3} on all 122 languages of the Belebele dataset \citep{bandarkar-etal-2024-belebele} described earlier.
We construct a zero-shot chain-of-thought (CoT) \citep{wei2022chain} prompt for each language.
Details of the construction pipeline and per-language curation are in Appendix~\ref{appsec: experimental-setups}.
The experiments are conducted on a single L40S GPU with the batch size set to 256.

\paragraph{Results.}

\begin{figure*}[t]
  \centering\includegraphics[width=0.8\textwidth]{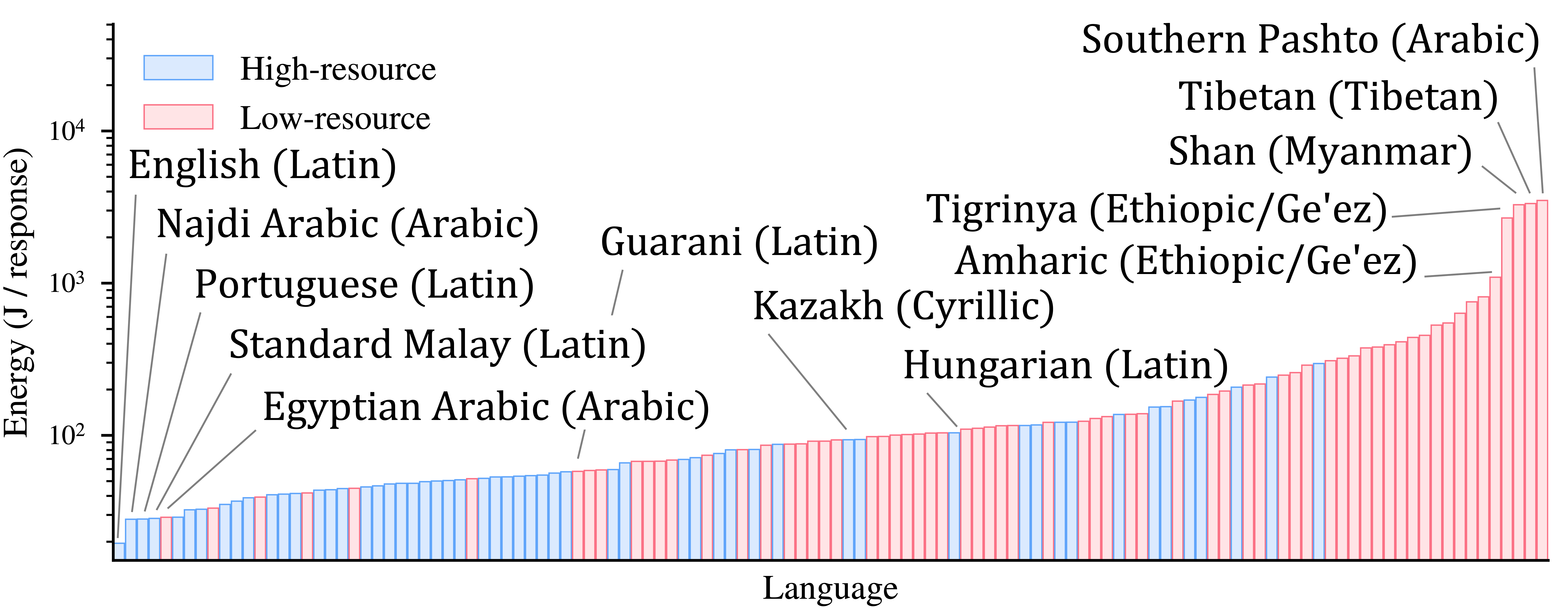}
  \caption{Energy per response across 122 languages (Qwen3-8B, zero-shot CoT, Belebele, $\log$-scale).
  Per-response energy compounds the per-token cost (\Cref{fig:energy_per_token}) with output-token count, capturing the end-to-end energy a user pays per query.
  The disparity spans $19.5$~J/response (English) to $3{,}496$~J/response (Pashto), a $179\times$ gap.
  }
  \label{fig:energy_per_response}
\end{figure*}

\Cref{fig:energy_per_response} presents the energy per response.
English uses $19.5$~J/response, whereas Pashto consumes $3{,}496$~J/response, Tibetan $3{,}337$~J, Shan $3{,}276$~J, and Tigrinya $2{,}681$~J, presenting a $179\times$ gap between the cheapest and most expensive languages.
Generally, low-resource languages consume much more energy per request than high-resource languages such as English.
There are notable exceptions, however: Egyptian Arabic (\texttt{arz\_Arab}), a low-resource language, sits within the cheapest five in both the per-response and per-token views.
The reason is because Egyptian Arabic shares its Arabic script with Modern Standard Arabic, a high-resource language with strong tokenizer coverage, and the cost benefit carries over to the low-resource variety.
The same pattern holds for other low-resource Arabic varieties (Mesopotamian, Najdi, Moroccan, North Levantine), all of which fall in the cheap cluster.

The energy  disparity is the \emph{product} of two factors, the number of output tokens generated per request, and the energy consumed per token.
Prior work has observed that low-resource languages elicit longer generations \citep{ahia-etal-2023-languages,petrov2023language}, and \Cref{tab:energy_decomposition} reaffirms this that the four low-resource languages in our subset emit $20$--$25\times$ more output tokens per response than English.

Beyond token count, however, our measurements reveal a second penalty: \textit{the per-token energy also varies substantially across languages}, by up to $8.3\times$.
\Cref{fig:energy_per_token} plots per-token energy for all 122 languages in Belebele, ordered by value and colored by resource level.
The distribution is heavily right-skewed: most languages cluster between $0.10$ and $0.20$~J/token, with a long tail of low-resource languages in non-Latin scripts (Tibetan, Pashto, Tigrinya, Shan, Amharic) exceeding $0.4$~J/token.\footnote{The complete per-language values (energy per token, output tokens, total energy, and accuracy) for all 122 languages are provided in Appendix~\ref{sec:appendix_languages}.}
The two penalties compound: the same five languages occupy the top of both \Cref{fig:energy_per_token} and \Cref{fig:energy_per_response}, so a per-token cost that is $\approx 7\times$ English combines with a token-count that is $\approx 20$--$25\times$ English to yield a per-response cost that is $\approx 170$--$179\times$ English.
This compounding is not coincidental as these languages consume more memory per response (the amount of KV cache is proportional to the number of tokens in the sequence).
Since serving systems are constrained by GPU memory, this forces servers serving such languages to run at smaller batch sizes that underutilize the GPU's compute units and therefore consume more energy per token~\citep{chung2026where}.\footnote{Smaller batch sizes underutilize the hardware's compute units and poorly amortize hardware overheads like static power draw, leading to higher energy consumption per FLOP.}

\begin{figure}[t]
  \includegraphics[width=\columnwidth]{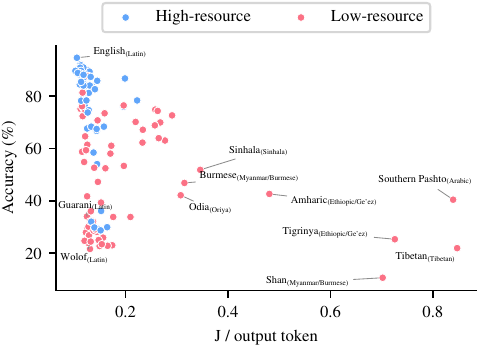}
  \caption{Output energy per token vs. accuracy across 122 languages (Qwen3-8B, zero-shot CoT, Belebele).
  The languages (typically the low-resource languages) on which the model has the highest energy per token are the ones on which it is least likely to answer correctly.}
  \label{fig:acc_vs_jpt}
\end{figure}

Moreover, \textit{low-resource languages are doubly penalized in performance and energy.}
\Cref{fig:acc_vs_jpt} plots accuracy against per-token energy for all 122 languages.
English achieves 94.6\% accuracy at the lowest energy, while the highest-energy languages (Tibetan, Pashto, Tigrinya, Shan, Amharic) all fall at or below 43\% accuracy, with Shan dropping to 10.6\%.
This is a \textit{double cost + performance penalty, the model both fails to answer correctly and consumes far more energy on these languages}.
These languages all use non-Latin scripts (Tibetan, Myanmar, Arabic, Ethiopic) with high character-to-token ratios.
For these languages, the model emits long, repetitive output that combines elevated per-token cost with inflated token counts.
We emphasize that this is a co-occurrence rather than a causal relationship: high per-token energy does not cause low accuracy, nor vice versa.
Both are downstream symptoms of the same upstream factors, tokenizer under-coverage and sparse pretraining representation for these languages, which simultaneously inflate token counts, and thus energy, and degrade model quality.

\begin{table*}[t]
  \centering
    \includegraphics[width=\linewidth]{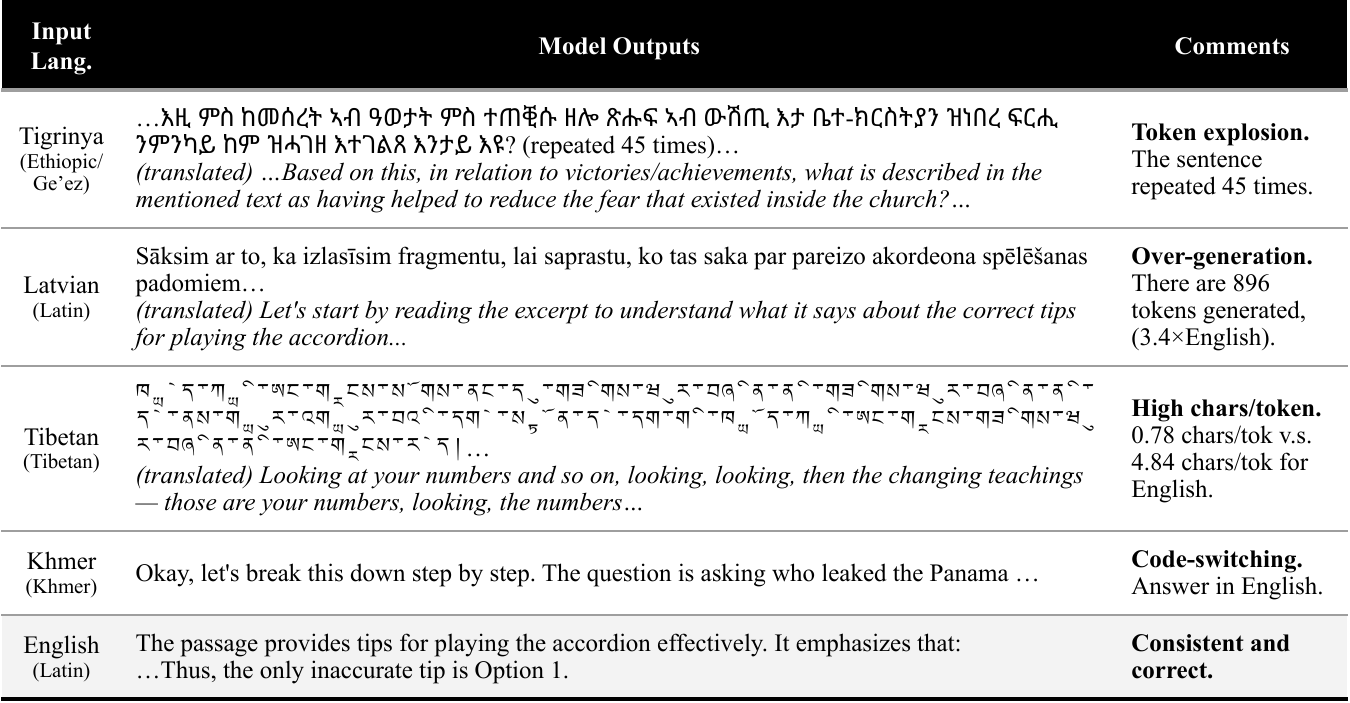}
  \caption{Representative Qwen3-8B generations on Belebele.
  We present common phenomena of the model on response generation; for contrast, we also show high-quality English outputs on the same task.
  }
  \label{tab:failure_modes}
\end{table*}
\Cref{tab:failure_modes} presents four representative generations alongside the English output for the same prompt.
We identify four recurring phenomena:
\begin{description}[leftmargin=1em,itemsep=2pt,topsep=2pt]
  \item[\textit{Token explosion.}] For Shan, Pashto, Tibetan, Tigrinya, Burmese, and Amharic, the model emits up to ${\sim}4{,}700$ output tokens per question, entering a repetitive loop \citep{Holtzman2020The} unable to produce a short answer.
  This regime is the most energy-wasteful: both the per-token cost and the token count are elevated.
  \item[\textit{Over-generation.}] Latvian, Lithuanian, Tswana, and Wolof generate more tokens than English at a similar per-token cost, suggesting the model partially processes the language but responds at length, reflecting uncertainty or partial understanding.
  \item[\textit{High character-to-token ratio.}] Scripts such as Myanmar (Burmese, Shan), Tibetan, Ethiopic (Amharic, Tigrinya), and Indic scripts (Gujarati, Kannada, Odia, Malayalam) require multiple Unicode characters per syllable and are tokenized inefficiently by models trained primarily on Latin-script text~\citep{ahia-etal-2023-languages,petrov2023language}. By contrast, Simplified Chinese (zho\_Hans) matches English at $\approx$0.10~J/token despite its non-Latin script, because modern Chinese is tokenized efficiently.
  \item[\textit{Code-switching.}] For some low-resource languages the model responds in English or another high-resource language \citep{winata-etal-2021-multilingual}.
\end{description}

We note that the energy disparity reflects both modeling choices and intrinsic linguistic structure.
On the modeling side, tokenizers trained on Latin-dominated corpora allocate few merges to underrepresented scripts \citep{petrov2023language}.
On the linguistic side, scripts and morphologies differ in information density:
abugidas like Ethiopic and Tibetan encode syllables across multiple Unicode
codepoints that subword tokenizers fragment inefficiently
\citep{ahia-etal-2023-languages, velayuthan-sarveswaran-2025-egalitarian};
agglutinative morphology (Finnish, Turkish, Korean) packs root, tense, case,
and agreement into single long words that BPE struggles to segment along
morpheme boundaries \citep{rust-etal-2021-good, toraman2023impact};
and abjads like Arabic omit short vowels that the model must infer from
context \citep{habash2010introduction}.
Chinese is a counterexample: zho\_Hans uses a non-Latin script but matches English at 0.10 J/token, showing that sufficient training-data coverage \citep{yang2025qwen3} can offset typological complexity.

\subsection{Energy Cost Disparity Persists Across Models}
\label{sec:cross_model}

\paragraph{Setup.}
We compare five models spanning three families and a range of sizes, Qwen3-8B, Qwen3-14B, and Qwen3-32B \citep{yang2025qwen3}, gemma-3-27B \citep{gemmateam2025gemma3technicalreport}, and Llama-3.1-8B-Instruct \citep{grattafiori2024llama}, on the 8-language subset specified in \S~\ref{sec:preliminaries}.
All models are served on L40S GPUs with the batch size set to 256.

\begin{table}[t]
\centering
\renewcommand{\arraystretch}{1.4}
\resizebox{\linewidth}{!}{
\begin{tabular}{clccccc}
\toprule
\multirow{2}{*}{\textbf{Res.}} & \multirow{2}{*}{\textbf{Lang.}} & \multicolumn{3}{c}{\textbf{Qwen3}} & \textbf{Gemma3} & \textbf{Llama3.1} \\
\cmidrule(lr){3-5} \cmidrule(lr){6-6} \cmidrule(lr){7-7}
 &  & \textbf{8B} & \textbf{14B} & \textbf{32B} & \textbf{27B} & \textbf{8B} \\
\midrule
                       & English\textsubscript{(Latin)} & \cellcolor[HTML]{FFFFFF}0.10 & \cellcolor[HTML]{FFFEFD}0.17 & \cellcolor[HTML]{FEF5F5}0.51 & \cellcolor[HTML]{FDF3F2}0.59 & \cellcolor[HTML]{FFFFFF}0.09 \\
                       & Chinese\textsubscript{(Simplified)} & \cellcolor[HTML]{FFFFFF}0.10 & \cellcolor[HTML]{FFFEFE}0.17 & \cellcolor[HTML]{FDF4F3}0.56 & \cellcolor[HTML]{FCF1F0}0.67 & \cellcolor[HTML]{FFFFFF}0.10 \\
                       & Russian\textsubscript{(Cyrillic)} & \cellcolor[HTML]{FFFFFF}0.12 & \cellcolor[HTML]{FFFDFD}0.20 & \cellcolor[HTML]{FDF1F0}0.70 & \cellcolor[HTML]{FCF0EF}0.72 & \cellcolor[HTML]{FFFFFF}0.10 \\
\multirow{-4}{*}{High} & French\textsubscript{(Latin)} & \cellcolor[HTML]{FFFFFF}0.11 & \cellcolor[HTML]{FFFDFD}0.19 & \cellcolor[HTML]{FDF3F2}0.60 & \cellcolor[HTML]{FCF1F0}0.69 & \cellcolor[HTML]{FFFFFF}0.10 \\
\midrule
                       & Southern Pashto\textsubscript{(Arabic)} & \cellcolor[HTML]{FCEDEC}0.84 & \cellcolor[HTML]{FAE1DF}1.36 & \cellcolor[HTML]{FBE9E8}1.00 & \cellcolor[HTML]{FCECEB}0.88 & \cellcolor[HTML]{FEF7F7}0.44 \\
                       & Tigrinya\textsubscript{(Ethiopic/Ge'ez)} & \cellcolor[HTML]{FDF0EF}0.73 & \cellcolor[HTML]{FBE8E6}1.06 & \cellcolor[HTML]{F9DFDD}1.43 & \cellcolor[HTML]{FBE8E7}1.04 & \cellcolor[HTML]{FFFCFC}0.24 \\
                       & Shan\textsubscript{(Myanmar/Burmese)} & \cellcolor[HTML]{FDF1F0}0.70 & \cellcolor[HTML]{F8DAD7}1.64 & \cellcolor[HTML]{F3BDB9}2.83 & \cellcolor[HTML]{EA8E86}4.79 & \cellcolor[HTML]{FDF2F1}0.66 \\
\multirow{-4}{*}{Low}  & Tibetan\textsubscript{(Tibetan)} & \cellcolor[HTML]{FCEDEC}0.85 & \cellcolor[HTML]{F7D1CE}2.02 & \cellcolor[HTML]{EB928B}4.60 & \cellcolor[HTML]{F8DAD8}1.63 & \cellcolor[HTML]{FEF5F5}0.52 \\
\bottomrule
\end{tabular}}
\caption{Per-language energy per output token (J / output token) across five models.
The cross-language disparity reproduces on every model, and the high- and low-resource groups never overlap on any model.}
\label{tab:cross-model-energy-belebele}
\end{table}

\paragraph{Results.}
\Cref{tab:cross-model-energy-belebele} reports per-token energy for all five models.
\textit{The cross-language disparity is present at every model size and in all three families.}
The ratio between the most and least expensive of the 8 languages is $8.3\times$ for Qwen3-8B, $12.2\times$ for Qwen3-14B, $9.0\times$ for Qwen3-32B, $8.2\times$ for gemma-3-27B, and $7.1\times$ for Llama-3.1-8B-Instruct.
English or Chinese is the cheapest language, and Tibetan or Shan is the most expensive on every model, and the high- and low-resource groups never overlap.
The absolute per-token energy rises with model size (e.g.\ English from $0.10$ J / token on Qwen3-8B to $0.17$ on Qwen3-14B and $0.51$ on Qwen3-32B).
However, the divide does not diminish with scale.
Across an 8B-to-32B span for Qwen 3 models, the four low-resource languages remain several times more expensive per token than the four high-resource ones at every size.

\subsection{Energy Cost Disparity Persists Across Inference Configurations}
\label{sec:cross_gpu}

\begin{figure*}[t]
  \centering
  \begin{minipage}{.9\linewidth}
    \centering
    {\includegraphics[width=0.7\textwidth]{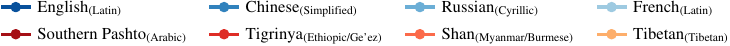}}
  \end{minipage}
  \begin{minipage}{.46\linewidth}
    \centering
    {\includegraphics[width=\textwidth]{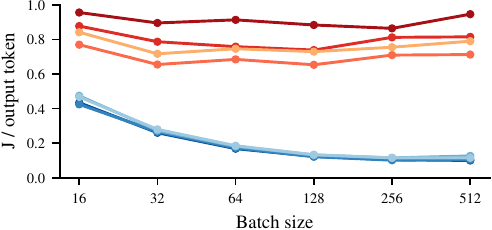}}
  \vspace{-0.5cm}
    \caption*{(a) L40S.}
  \end{minipage}
  \quad
  \begin{minipage}{.46\linewidth}
    \centering
    {\includegraphics[width=\textwidth]{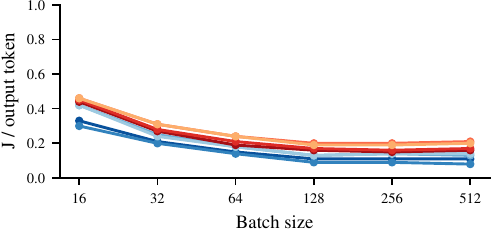}}
  \vspace{-0.5cm}
    \caption*{(b) RTX 6000 Pro Blackwell.}
  \end{minipage}
  \vspace{-0.2cm}
  \caption{Per token energy v.s. batch size of Qwen3-8B on L40S and RTX~6000 across 8 languages on Belebele.}
  \label{fig:cross-gpu}
\end{figure*}

\paragraph{Setup.}
We provide studies into two GPU types (NVIDIA L40S of 48G GPU memory and RTX~6000 Pro Blackwell of 96G GPU memory) and batch sizes ($\in\{16,32,64,128,256,512\}$) for the inference configurations.
We run Qwen3-8B and use the same 8-language subset (four high-resource and four low-resource) as \S~\ref{sec:cross_model}.

\paragraph{Results.}
In \Cref{fig:cross-gpu}, we observe that the \textit{per-token energy differs across GPUs, but the energy disparity across languages persists at every batch size.}
For instance, simplified Chinese remains one of the cheapest languages at every batch size, while Shan is the most expensive at most batch sizes.
The per-token energy disparity stems primarily from sequence-length effects.
Low-resource languages produce longer token sequences, which inflate KV cache size and require attention and other computations over longer contexts, both during prefill (generating the first output token) and decoding, increasing the amount of computation and memory accesses.
As batch size grows, memory pressure rises, and these effects compound across requests.

\subsection{Energy Cost Disparity Persists Across Tasks}
\label{sec:lmarena}

\paragraph{Setup.}
We now test how the task would influence the energy cost disparity.
We use three tasks that span the distinct usage scenarios end users have with LLMs, and the corresponding datasets described in \S~\ref{sec:preliminaries}: knowledge-grounded question answering (Belebele \citep{bandarkar-etal-2024-belebele}), structured multi-step reasoning (GSM8K \citep{cobbe2021gsm8k}), and open-ended conversational assistance (LM-Arena \citep{chiang2024chatbot}).
We hold the model (Qwen3-8B), hardware (a single L40S), and serving configuration (batch size of 256) fixed on the same 8-language subset as \S~\ref{sec:cross_model}.

\begin{table}[t]
  \centering
  \small
  \renewcommand{\arraystretch}{1.4}
  \resizebox{\linewidth}{!}{
  \begin{tabular}{clrrr}
    \toprule
    & & \multicolumn{3}{c}{\textbf{J\,/\,output token}} \\
    \cmidrule(lr){3-5}
    \textbf{Resource} & \textbf{Language} & \textbf{Belebele} & \textbf{GSM8K} & \textbf{LM-A} \\
    & & {(reading)} & {\footnotesize(math)} & {\footnotesize(chat)} \\
    \midrule
    \multirow{4}{*}{High} & English\textsubscript{(Latin)} & \cellcolor[HTML]{FFFBFB}0.105 & \cellcolor[HTML]{FFFCFC}0.086 & \cellcolor[HTML]{FFFBFB}0.097 \\
     & Chinese\textsubscript{(Simplified)} & \cellcolor[HTML]{FFFBFB}0.102 & \cellcolor[HTML]{FFFCFC}0.087 & \cellcolor[HTML]{FFFBFB}0.098 \\
     & Russian\textsubscript{(Cyrillic)} & \cellcolor[HTML]{FFFAFA}0.118 & \cellcolor[HTML]{FFFCFC}0.089 & \cellcolor[HTML]{FFFBFB}0.105 \\
     & French\textsubscript{(Latin)} & \cellcolor[HTML]{FFFAFA}0.113 & \cellcolor[HTML]{FFFCFC}0.091 & \cellcolor[HTML]{FFFBFA}0.107 \\
    \midrule
    \multirow{4}{*}{Low} & Southern Pashto\textsubscript{(Arabic)} & \cellcolor[HTML]{F4C6C2}0.840 & \cellcolor[HTML]{FFFAFA}0.117 & \cellcolor[HTML]{F3BDB9}0.957 \\
     & Tigrinya\textsubscript{(Ethiopic/Ge'ez)} & \cellcolor[HTML]{F6CECA}0.726 & \cellcolor[HTML]{F4C3BF}0.872 & \cellcolor[HTML]{F3BEB9}0.946 \\
     & Shan\textsubscript{(Myanmar/Burmese)} & \cellcolor[HTML]{F6D0CC}0.702 & \cellcolor[HTML]{F9DDDB}0.516 & \cellcolor[HTML]{F8D8D5}0.593 \\
     & Tibetan\textsubscript{(Tibetan)} & \cellcolor[HTML]{F4C5C1}0.847 & \cellcolor[HTML]{F2BBB6}0.990 & \cellcolor[HTML]{F2B9B3}1.023 \\
    \bottomrule
  \end{tabular}}
  \caption{Energy per output token for Qwen3-8B on the same 8-language subset across three datasets.
  The cross-language disparity persists under every task, with English or Chinese being cheapest and Tibetan most expensive in all three.}
  \label{tab:cross_task}
\end{table}

\paragraph{Results.}
\Cref{tab:cross_task} reports energy per output token under all three tasks.
\textit{The cross-language disparity is present in every task.}
The ratio between the most and least expensive of the 8 languages is $8.3\times$ for Belebele (reading comprehension), $11.6\times$ for GSM8K (math), and $10.6\times$ for LM-Arena (chat).
English or Chinese is the cheapest, and Tibetan is the most expensive language in all three tasks.
The high- vs. low-resource gap persists across all tasks, with the energy cost falling disproportionately on the four low-resource, non-Latin-script languages.

\section{Actionable Items and Recommendations}
\label{sec: recommendations}

\paragraph{Treat energy as a first-class evaluation axis.}
We advocate for \textit{reporting energy consumption as a standard companion to existing performance metrics across NLP research}. Our findings clearly indicate that
performance metrics such as accuracy alone are an incomplete picture of what a system delivers in deployment. For instance, two systems that achieve comparable accuracy at substantially different energy costs are not equivalent, and our evaluation practices should reflect that.

We propose two concrete venues for energy reporting.
First, extending the Responsible NLP Checklist \citep{rogers-etal-2021-just-think}, to also include
energy consumption as a standard checklist item, so that energy cost becomes visible alongside existing performance-centered metrics. This is by and large feasible,  as energy is a measurable physical quantity that can be collected with existing tooling on the same hardware used for evaluation.
Second, model developers should measure and report inference energy alongside the existing metrics in model cards.
We note that training carbon emission is already an optional field of the model card spec \citep{luccioni2023estimating, faiz2024llmcarbon}, but it obscures the amount of energy consumption by multiplying it with estimated carbon intensity; we believe  the measurable physical quantity of energy will serve as a more concrete reporting metric.

\paragraph{Multilingual models should report energy consumption for each individual language.} A model may incur unequal energy costs when serving different users, and these costs should be explicitly reported.
A model that meets a 95\% accuracy threshold on English at 0.10~J/token and 22\% on Tibetan at 0.85~J/token represents, in deployment terms, two qualitatively different systems for two communities of users: cheap and reliable for one, expensive and broken for the other.
Pooling these into a single ``multilingual model'' claim hides both the failure mode and the cost mode.

We argue that any ``multilingual'' coverage claims in models and papers should be  backed by {\it per-language energy reports} in papers, model cards, and on leaderboards. Further, the factors that drive disproportionate energy consumption in certain languages should be investigated (e.g., through ablation studies),  rather than focusing solely on accuracy improvements.
Re-framing evaluation in these terms changes which models are considered to have ``succeeded'' on multilingual coverage, and reorients the incentives that drive multilingual model development.

\paragraph{Language-aware serving optimization.}
Existing serving optimizations are unaware of the \emph{language mixture} of workloads and choose configurations based on the \emph{average} language of the expected incoming workload~\citep{distserve-osdi24,sarathiserve-osdi24,nanoflow-osdi25,cornfigurator-arxiv26}.
However, as we have shown, language is a high-impact factor; it has broad serving implications, including the number of tokens, energy consumption, and so on.
This means that when workloads are segmented into separate languages, each language may have different optimal configurations.
For instance, different languages are read and spoken at different speeds~\citep{reading-speed-jml19,speaking-speed-languages}.
In interactive applications like ChatGPT, languages that are read or spoken by native users at slower speeds (tokens per second) have longer time budgets to produce each output token, which allows the system to use larger batch sizes that consume less energy per token.
Deeper and broader understanding of various languages (e.g., reading speed) will inform such optimizations and lead to higher efficiency.

Inference systems can also curb energy waste at serving time by adapting decoding constraints to the input.
For multilingual deployments, we may use language detection to apply per-language maximum output lengths.
We can leverage efforts from the generation community and apply adaptive output caps \citep{takase-okazaki-2019-positional}, early-stopping criteria \citep{wang2026estar}, repetition penalties \citep{welleck-etal-2020-consistency}, and so on, to ensure better output control.

\paragraph{Language-aware query routing.}
Rather than serving every input with a single generalist model, system designers can route queries based on the input language.
One strategy is to route queries to smaller, specialized models when the input profile permits it \citep{ong2025routellm, chen2024frugalgpt}.
In the multilingual case, this means sending low-resource language queries to language-specialized models that are smaller and more efficient for their target language.
Routing infrastructure already exists in production systems for cost reasons, and extending it to account for energy is a low-friction change with substantial aggregate impact.

A complementary strategy is to route eligible queries through a translate-then-process pipeline: translating non-English inputs to English, generating in English, and translating the output back to the source language.
Because per-token energy costs in English are substantially lower and English generations require fewer tokens, this pipeline can reduce total energy consumption even after accounting for the overhead of two translation steps.
Prior work has shown that such translate-then-process pipelines can also \emph{improve} task accuracy for low-resource languages, since the core reasoning step occurs in the language where the model is good at \citep{shi2023language, chen-etal-2023-revisiting,kowtal-etal-2024-chain, ko-etal-2025-understand}.
Energy efficiency and task performance, in these cases, can both be improved.
We note however that this strategy may not be applicable in the case of tasks that hinge on cultural nuance, language-specific commonsense, idiomatic expression, or sociolinguistic register, where a translated output may be fluent but not culturally faithful  \citep{liu-etal-2025-translation}.

\paragraph{Treat per-language disparity as a deployment equity issue.}
The energy disparities translate directly into deployment-level inequities.
Since energy costs are a primary driver of inference pricing, providers may, deliberately or not, impose higher costs on users of low-resource languages or exclude those languages from service to manage costs.
Meanwhile, the aggregate environmental cost falls on all users, while the model failures fall disproportionately on speakers of underserved languages.
We therefore believe that providers, alongside the research community, should treat per-language energy disparity as a deployment-time equity issue and to develop concrete mechanisms for addressing it, such as equity audits at model release, transparent per-language pricing, or subsidized inference for underserved languages.

\section{Conclusion}

We presented the first systematic measurement of inference energy consumption across languages in LLM serving, spanning 122 languages, alongside various models, inference configurations, and three tasks.
Our main finding is a stark and persistent \emph{language--energy divide}: per-token energy varies by up to $8.3\times$ across languages on a single model, and total energy per response varies by up to $179\times$ between English and Pashto.
The divide persists across models, GPUs, batch sizes, and tasks, indicating a systemic property of multilingual LLM deployment.
Worse, the languages most expensive to serve are also those on which models perform worst, imposing a double penalty on speakers of underserved languages.
Our analysis revealed that the disparity is driven by two factors: low-resource and non-Latin-script languages incur both inflated per-token energy costs and longer output sequences.
We argue that the community should treat energy as a first-class evaluation axis, report per-language energy in checklists and model cards, and adopt and innovate deployment-side mitigation methods.
Closing the language--energy divide is not only an efficiency problem but an equity problem, and addressing it is essential if multilingual LLMs are to serve all their users fairly.

\section*{Acknowledgement}
We thank the anonymous reviewers for their feedback on this work.
This work was supported in part by NSF grants CCF-2450085 and CNS-2106184, DARPA ML2P Award HR0011-26-9-E190, and by grants from Ford, Laude Institute, OpenAI, and the Survival and Flourishing Fund.
Jae-Won Chung was additionally supported by the Kwanjeong Educational Foundation and the Rackham Predoctoral Fellowship.
Yulong Chen was supported by the DARPA program SciFy.
Any opinions, findings, conclusions or recommendations expressed in this material are those of the authors and do not necessarily reflect the views of the National Science Foundation, Ford, Laude Institute, Open AI or the Survival and Flourishing Fund.

\section*{Limitations}

We discuss the limitations of our work here.
First, due to resource limitations, our cross-hardware comparison is limited to NVIDIA L40S and RTX~6000 Pro Blackwell GPUs.
We highlight that absolute per-token energy values are hardware-specific, and the magnitude of the disparity can differ across GPUs.
Our findings should be re-validated as serving hardware evolves.
Second, we do not evaluate commercial closed-source models such as GPT, Gemini, or Claude, which would require API-based energy estimation or access to provider infrastructure.
Given that these models serve a substantial share of multilingual queries in deployment, extending our methodology to them is an important direction for future work, but it requires support from the companies.
Third, Belebele and our translated GSM8K and LM-Arena prompts are standardized translations rather than natively authored text in each language.
They may not fully reflect how each language is used in practice, particularly for orthography, register, and dialectal variation.
The cross-language energy gap we measure is therefore best interpreted as a lower bound on the disparity faced by users producing natively authored, often noisier, real-world inputs.

\section*{Ethical Considerations}

This work is fundamentally motivated by equity concerns, and we discuss them here.

\paragraph{Risk of language deprioritization.}
A central finding of this paper is that low-resource languages cost more energy to serve than high-resource ones.
We are aware that publishing this result carries a dual-use risk: cost-conscious providers could use it to justify deprioritizing, rate-limiting, or excluding low-resource languages from service.
We believe the alternative, leaving these disparities unmeasured and undiscussed, is worse, because it allows the same outcomes to occur silently and without accountability.
We have therefore framed our recommendations around \emph{transparency} (per-language energy reporting), \emph{mitigation} (language-aware serving, routing, and translate-then-process pipelines), and \emph{equity audits at deployment}.
We urge providers and researchers using our findings to do the same.

\paragraph{Whose costs, whose benefits.}
The energy costs we document are paid in aggregate, through electricity, water, and carbon, by all users and ultimately by the broader public, while the worst service quality is borne disproportionately by speakers of underserved languages.
This asymmetry between who pays and who is served is itself an ethical issue, and one that performance-only evaluation has rendered invisible.
We hope our work contributes to making it visible.

\paragraph{Environmental impact of this study.}
Our measurements required running LLM inference across many languages, models, hardware, and inference configurations, which itself consumed non-trivial energy.
We view this as a one-time cost in service of changing community practice.
To amortize it, we will release all per-language energy measurements, prompts, and processing scripts so that downstream researchers can build on our numbers rather than re-running the full sweep.

\paragraph{Translation quality and representation.}
Our translated prompts for GSM8K and LM-Arena, and the per-language instruction and primer for Belebele, were produced via NLLB-200 with back-translation quality control, and hand-curated where the automated pipeline failed. Despite these checks, machine-translated and even hand-curated prompts cannot fully substitute for native authorship by speakers of each language, and our results should not be taken as a verdict on the intrinsic difficulty or value of any language.
They are a verdict on how current LLM stacks treat each language under current tokenization, training data, and serving choices.

\paragraph{Recommendations and downstream effects.}
We recommend, among other interventions, translate-then-process pipelines for tasks where the generation is largely language-agnostic.
We note that uncritical application of such pipelines to culturally-loaded or sociolinguistically-rich tasks risks producing outputs that are fluent but culturally unfaithful, and we explicitly caution against this in \S~\ref{sec: recommendations}.
Energy efficiency should not become a justification for erasing language-specific meaning.

\bibliography{custom}

\appendix

\onecolumn
\begingroup
\footnotesize
\setlength{\tabcolsep}{4.5pt}
\renewcommand{\arraystretch}{1.4}
\begin{longtable}{@{}llrrrrr@{}}
  \toprule
  \textbf{Language} & \textbf{Code} & \textbf{Resource} & \textbf{J/tok} & \textbf{Output Tokens} & \textbf{Total Energy (J)} & \textbf{Accuracy} \\
  \midrule
  \endfirsthead

  \multicolumn{7}{c}{\tablename\ \thetable\ -- \textit{Continued from previous page}} \\
  \toprule
  \textbf{Language} & \textbf{Code} & \textbf{Resource} & \textbf{J/tok} & \textbf{Output Tokens} & \textbf{Total Energy (J)} & \textbf{Accuracy} \\
  \midrule
  \endhead

  \midrule
  \multicolumn{7}{r}{\textit{Continued on next page}} \\
  \endfoot

  \bottomrule
  \endlastfoot
    Southern Pashto                & pbt\_Arab & {\color[HTML]{FB7185} Low}  & \cellcolor[HTML]{EB958D}0.84 & \cellcolor[HTML]{EEA29B}3747646 & \cellcolor[HTML]{EB938B}3146541.2 & \cellcolor[HTML]{F5CDCA}40.4 \\
Tibetan                        & bod\_Tibt & {\color[HTML]{FB7185} Low}  & \cellcolor[HTML]{EB938C}0.85 & \cellcolor[HTML]{EFA8A2}3543429 & \cellcolor[HTML]{EC9890}3002866.9 & \cellcolor[HTML]{EFADA7}21.9 \\
Shan                           & shn\_Mymr & {\color[HTML]{FB7185} Low}  & \cellcolor[HTML]{EFAAA4}0.70 & \cellcolor[HTML]{EB948D}4200313 & \cellcolor[HTML]{EC9A93}2948722.2 & \cellcolor[HTML]{EB9992}10.6 \\
Tigrinya                       & tir\_Ethi & {\color[HTML]{FB7185} Low}  & \cellcolor[HTML]{EEA59F}0.73 & \cellcolor[HTML]{F0AEA9}3325414 & \cellcolor[HTML]{F0ACA7}2413327.9 & \cellcolor[HTML]{F0B3AE}25.3 \\
Amharic                        & amh\_Ethi & {\color[HTML]{FB7185} Low}  & \cellcolor[HTML]{F5CBC7}0.48 & \cellcolor[HTML]{F7D4D1}2048079 & \cellcolor[HTML]{F9E0DD}983077.9  & \cellcolor[HTML]{F6D1CE}42.6 \\
Sinhala                        & sin\_Sinh & {\color[HTML]{FB7185} Low}  & \cellcolor[HTML]{F9DEDC}0.35 & \cellcolor[HTML]{F7D2CF}2116100 & \cellcolor[HTML]{FBE8E7}740635.0  & \cellcolor[HTML]{F9E1DF}51.8 \\
Burmese                        & mya\_Mymr & {\color[HTML]{FB7185} Low}  & \cellcolor[HTML]{FAE4E3}0.31 & \cellcolor[HTML]{F7D1CE}2154191 & \cellcolor[HTML]{FCEBE9}667799.2  & \cellcolor[HTML]{F7D9D6}46.8 \\
Kannada                        & kan\_Knda & {\color[HTML]{FB7185} Low}  & \cellcolor[HTML]{FBE7E6}0.29 & \cellcolor[HTML]{F8D7D4}1963275 & \cellcolor[HTML]{FCEEED}569349.8  & \cellcolor[HTML]{EFF9F4}72.6 \\
Khmer                          & khm\_Khmr & {\color[HTML]{FB7185} Low}  & \cellcolor[HTML]{FBE9E7}0.28 & \cellcolor[HTML]{F9DCDA}1778891 & \cellcolor[HTML]{FDF1F0}498089.5  & \cellcolor[HTML]{FDF5F5}63.0 \\
Gujarati                       & guj\_Gujr & {\color[HTML]{FB7185} Low}  & \cellcolor[HTML]{FBEAE9}0.27 & \cellcolor[HTML]{F9DCDA}1781503 & \cellcolor[HTML]{FDF1F0}481005.8  & \cellcolor[HTML]{F9FDFB}69.9 \\
Eastern Panjabi                & pan\_Guru & {\color[HTML]{FB7185} Low}  & \cellcolor[HTML]{FCECEB}0.26 & \cellcolor[HTML]{FAE3E1}1556173 & \cellcolor[HTML]{FDF4F3}404605.0  & \cellcolor[HTML]{E8F6EF}74.3 \\
Tamil                          & tam\_Taml & {\color[HTML]{FB7185} Low}  & \cellcolor[HTML]{FCECEB}0.26 & \cellcolor[HTML]{FAE3E1}1540155 & \cellcolor[HTML]{FDF4F4}400440.3  & \cellcolor[HTML]{FEFFFE}68.8 \\
Odia                           & ory\_Orya & {\color[HTML]{FB7185} Low}  & \cellcolor[HTML]{FAE4E3}0.31 & \cellcolor[HTML]{FCEDEC}1204527 & \cellcolor[HTML]{FEF5F5}373403.4  & \cellcolor[HTML]{F6D0CD}42.1 \\
Telugu                         & tel\_Telu & {\color[HTML]{FB7185} Low}  & \cellcolor[HTML]{FCECEB}0.26 & \cellcolor[HTML]{FBE9E8}1339006 & \cellcolor[HTML]{FEF6F5}348141.6  & \cellcolor[HTML]{FDF7F6}63.9 \\
Lao                            & lao\_Laoo & {\color[HTML]{FB7185} Low}  & \cellcolor[HTML]{FDF0EF}0.23 & \cellcolor[HTML]{FBE6E4}1469386 & \cellcolor[HTML]{FEF6F6}337958.8  & \cellcolor[HTML]{FCF4F3}62.2 \\
Assamese                       & asm\_Beng & {\color[HTML]{FB7185} Low}  & \cellcolor[HTML]{FDF0EF}0.23 & \cellcolor[HTML]{FBE6E5}1445450 & \cellcolor[HTML]{FEF7F6}332453.5  & \cellcolor[HTML]{FEFCFC}67.1 \\
Malayalam                      & mal\_Mlym & {\color[HTML]{FB7185} Low}  & \cellcolor[HTML]{FCECEB}0.26 & \cellcolor[HTML]{FCEFEE}1161153 & \cellcolor[HTML]{FEF8F7}301899.8  & \cellcolor[HTML]{E6F5EE}74.8 \\
Sindhi                         & snd\_Arab & {\color[HTML]{FB7185} Low}  & \cellcolor[HTML]{FEF9F9}0.17 & \cellcolor[HTML]{F9DFDD}1678582 & \cellcolor[HTML]{FEF8F8}285358.9  & \cellcolor[HTML]{FCF2F1}61.0 \\
Armenian                       & hye\_Armn & {\color[HTML]{FB7185} Low}  & \cellcolor[HTML]{FDF2F1}0.22 & \cellcolor[HTML]{FCECEA}1267931 & \cellcolor[HTML]{FEF9F8}278944.8  & \cellcolor[HTML]{F8FDFB}70.1 \\
Bengali                        & ben\_Beng & {\color[HTML]{60A5FA} High} & \cellcolor[HTML]{FDF2F1}0.22 & \cellcolor[HTML]{FCEEEC}1199146 & \cellcolor[HTML]{FEF9F9}263812.1  & \cellcolor[HTML]{D9F0E4}78.3 \\
Nepali                         & npi\_Deva & {\color[HTML]{FB7185} Low}  & \cellcolor[HTML]{FDF3F3}0.21 & \cellcolor[HTML]{FCECEB}1242288 & \cellcolor[HTML]{FEF9F9}260880.5  & \cellcolor[HTML]{F3C2BE}33.8 \\
Swati                          & ssw\_Latn & {\color[HTML]{FB7185} Low}  & \cellcolor[HTML]{FFFBFB}0.16 & \cellcolor[HTML]{FBE7E6}1409901 & \cellcolor[HTML]{FFFAFA}225584.2  & \cellcolor[HTML]{EFAEA9}22.8 \\
Maltese                        & mlt\_Latn & {\color[HTML]{60A5FA} High} & \cellcolor[HTML]{FFFEFE}0.14 & \cellcolor[HTML]{FAE2E0}1581018 & \cellcolor[HTML]{FFFBFA}221342.5  & \cellcolor[HTML]{FBEDEC}58.4 \\
Igbo                           & ibo\_Latn & {\color[HTML]{FB7185} Low}  & \cellcolor[HTML]{FEF9F9}0.17 & \cellcolor[HTML]{FCEBEA}1286472 & \cellcolor[HTML]{FFFBFA}218700.2  & \cellcolor[HTML]{EFAFA9}23.0 \\
Tajik                          & tgk\_Cyrl & {\color[HTML]{FB7185} Low}  & \cellcolor[HTML]{FEF9F9}0.17 & \cellcolor[HTML]{FCEFEE}1149221 & \cellcolor[HTML]{FFFCFB}195367.6  & \cellcolor[HTML]{FBECEB}58.0 \\
Fulfulde                       & fuv\_Latn & {\color[HTML]{FB7185} Low}  & \cellcolor[HTML]{FFFFFF}0.13 & \cellcolor[HTML]{FBE5E4}1475561 & \cellcolor[HTML]{FFFCFB}191822.9  & \cellcolor[HTML]{EFACA7}21.6 \\
Zulu                           & zul\_Latn & {\color[HTML]{60A5FA} High} & \cellcolor[HTML]{FFFCFC}0.15 & \cellcolor[HTML]{FCEDEB}1232890 & \cellcolor[HTML]{FFFCFC}184933.5  & \cellcolor[HTML]{F1B9B4}28.7 \\
Marathi                        & mar\_Deva & {\color[HTML]{FB7185} Low}  & \cellcolor[HTML]{FDF5F4}0.20 & \cellcolor[HTML]{FEF7F6}896219  & \cellcolor[HTML]{FFFCFC}179243.8  & \cellcolor[HTML]{E0F3EA}76.4 \\
Bambara                        & bam\_Latn & {\color[HTML]{FB7185} Low}  & \cellcolor[HTML]{FFFCFC}0.15 & \cellcolor[HTML]{FDF0EF}1119522 & \cellcolor[HTML]{FFFDFC}167928.3  & \cellcolor[HTML]{EFAEA9}22.7 \\
Greek                          & ell\_Grek & {\color[HTML]{60A5FA} High} & \cellcolor[HTML]{FDF5F4}0.20 & \cellcolor[HTML]{FEF9F9}804328  & \cellcolor[HTML]{FFFDFD}160865.6  & \cellcolor[HTML]{B8E2CE}86.7 \\
Central Kurdish                & ckb\_Arab & {\color[HTML]{FB7185} Low}  & \cellcolor[HTML]{FEF8F7}0.18 & \cellcolor[HTML]{FEF8F7}859082  & \cellcolor[HTML]{FFFDFD}154634.8  & \cellcolor[HTML]{F3C2BE}33.8 \\
Bengali (Latin)                & ben\_Latn & {\color[HTML]{60A5FA} High} & \cellcolor[HTML]{FFFBFB}0.16 & \cellcolor[HTML]{FEF5F5}935815  & \cellcolor[HTML]{FFFDFD}149730.4  & \cellcolor[HTML]{F2BBB6}29.9 \\
Hindi                          & hin\_Deva & {\color[HTML]{60A5FA} High} & \cellcolor[HTML]{FDF5F4}0.20 & \cellcolor[HTML]{FFFCFC}708854  & \cellcolor[HTML]{FFFDFD}141770.8  & \cellcolor[HTML]{E2F4EB}75.9 \\
Tswana                         & tsn\_Latn & {\color[HTML]{60A5FA} High} & \cellcolor[HTML]{FFFCFC}0.15 & \cellcolor[HTML]{FEF5F5}940184  & \cellcolor[HTML]{FFFDFD}141027.6  & \cellcolor[HTML]{F1BAB5}29.2 \\
Nyanja                         & nya\_Latn & {\color[HTML]{FB7185} Low}  & \cellcolor[HTML]{FFFCFC}0.15 & \cellcolor[HTML]{FEF8F8}846511  & \cellcolor[HTML]{FFFEFE}126976.7  & \cellcolor[HTML]{F0B2AD}25.0 \\
Xhosa                          & xho\_Latn & {\color[HTML]{60A5FA} High} & \cellcolor[HTML]{FFFEFE}0.14 & \cellcolor[HTML]{FEF7F6}887370  & \cellcolor[HTML]{FFFEFE}124231.8  & \cellcolor[HTML]{F2BBB6}29.8 \\
Jingpho                        & kac\_Latn & {\color[HTML]{FB7185} Low}  & \cellcolor[HTML]{FFFBFB}0.16 & \cellcolor[HTML]{FFFAFA}765203  & \cellcolor[HTML]{FFFEFE}122432.5  & \cellcolor[HTML]{F0B4AF}25.9 \\
Nepali (Latin)                 & npi\_Latn & {\color[HTML]{FB7185} Low}  & \cellcolor[HTML]{FFFCFC}0.15 & \cellcolor[HTML]{FEF9F9}814114  & \cellcolor[HTML]{FFFEFE}122117.1  & \cellcolor[HTML]{F5CCC8}39.3 \\
Wolof                          & wol\_Latn & {\color[HTML]{FB7185} Low}  & \cellcolor[HTML]{FFFFFF}0.13 & \cellcolor[HTML]{FEF7F6}880761  & \cellcolor[HTML]{FFFEFE}114498.9  & \cellcolor[HTML]{F0B1AC}24.4 \\
Hausa                          & hau\_Latn & {\color[HTML]{FB7185} Low}  & \cellcolor[HTML]{FFFFFF}0.13 & \cellcolor[HTML]{FEF7F7}876624  & \cellcolor[HTML]{FFFEFE}113961.1  & \cellcolor[HTML]{F0B0AB}23.9 \\
Modern Standard Arabic (Latin) & arb\_Latn & {\color[HTML]{60A5FA} High} & \cellcolor[HTML]{FFFCFC}0.15 & \cellcolor[HTML]{FFFCFC}720450  & \cellcolor[HTML]{FFFFFF}108067.5  & \cellcolor[HTML]{F4C6C2}36.1 \\
Yoruba                         & yor\_Latn & {\color[HTML]{FB7185} Low}  & \cellcolor[HTML]{FFFCFC}0.15 & \cellcolor[HTML]{FFFCFC}713713  & \cellcolor[HTML]{FFFFFF}107057.0  & \cellcolor[HTML]{EFB0AA}23.4 \\
Swahili                        & swh\_Latn & {\color[HTML]{60A5FA} High} & \cellcolor[HTML]{FFFEFE}0.14 & \cellcolor[HTML]{FFFBFA}760846  & \cellcolor[HTML]{FFFFFF}106518.4  & \cellcolor[HTML]{FAE5E4}54.0 \\
Shona                          & sna\_Latn & {\color[HTML]{FB7185} Low}  & \cellcolor[HTML]{FFFEFE}0.14 & \cellcolor[HTML]{FFFBFA}756672  & \cellcolor[HTML]{FFFFFF}105934.1  & \cellcolor[HTML]{F1B9B4}28.6 \\
Tsonga                         & tso\_Latn & {\color[HTML]{FB7185} Low}  & \cellcolor[HTML]{FFFEFE}0.14 & \cellcolor[HTML]{FFFBFB}741420  & \cellcolor[HTML]{FFFFFF}103798.8  & \cellcolor[HTML]{F1BAB5}29.1 \\
Hindi (Latin)                  & hin\_Latn & {\color[HTML]{60A5FA} High} & \cellcolor[HTML]{FFFEFE}0.14 & \cellcolor[HTML]{FFFBFB}735045  & \cellcolor[HTML]{FFFFFF}102906.3  & \cellcolor[HTML]{FEFCFB}66.7 \\
Maori                          & mri\_Latn & {\color[HTML]{FB7185} Low}  & \cellcolor[HTML]{FFFEFE}0.14 & \cellcolor[HTML]{FFFBFB}734697  & \cellcolor[HTML]{FFFFFF}102857.6  & \cellcolor[HTML]{F2BFBA}32.1 \\
Southern Sotho                 & sot\_Latn & {\color[HTML]{60A5FA} High} & \cellcolor[HTML]{FFFFFF}0.13 & \cellcolor[HTML]{FEFAF9}785951  & \cellcolor[HTML]{FFFFFF}102173.6  & \cellcolor[HTML]{F2BFBA}32.0 \\
Halh Mongolian                 & khk\_Cyrl & {\color[HTML]{FB7185} Low}  & \cellcolor[HTML]{FFFBFB}0.16 & \cellcolor[HTML]{FFFEFE}634984  & \cellcolor[HTML]{FFFFFF}101597.4  & \cellcolor[HTML]{F9E3E1}52.4 \\
Kyrgyz                         & kir\_Cyrl & {\color[HTML]{FB7185} Low}  & \cellcolor[HTML]{FFFCFC}0.15 & \cellcolor[HTML]{FFFDFD}676822  & \cellcolor[HTML]{FFFFFF}101523.3  & \cellcolor[HTML]{F7D9D7}47.2 \\
Northern Sotho                 & nso\_Latn & {\color[HTML]{FB7185} Low}  & \cellcolor[HTML]{FFFFFF}0.13 & \cellcolor[HTML]{FFFBFB}730114  & \cellcolor[HTML]{FFFFFF}94914.8   & \cellcolor[HTML]{F2BBB7}30.1 \\
N. Azerbaijani                 & azj\_Latn & {\color[HTML]{FB7185} Low}  & \cellcolor[HTML]{FFFCFC}0.15 & \cellcolor[HTML]{FFFEFE}631317  & \cellcolor[HTML]{FFFFFF}94697.6   & \cellcolor[HTML]{F6FCF9}70.7 \\
Ganda                          & lug\_Latn & {\color[HTML]{FB7185} Low}  & \cellcolor[HTML]{FFFFFF}0.13 & \cellcolor[HTML]{FFFCFB}726949  & \cellcolor[HTML]{FFFFFF}94503.4   & \cellcolor[HTML]{F1B6B1}26.8 \\
Kinyarwanda                    & kin\_Latn & {\color[HTML]{FB7185} Low}  & \cellcolor[HTML]{FFFFFF}0.13 & \cellcolor[HTML]{FFFCFB}722516  & \cellcolor[HTML]{FFFFFF}93927.1   & \cellcolor[HTML]{F1BAB5}29.2 \\
Urdu (Latin)                   & urd\_Latn & {\color[HTML]{FB7185} Low}  & \cellcolor[HTML]{FFFEFE}0.14 & \cellcolor[HTML]{FFFEFE}656631  & \cellcolor[HTML]{FFFFFF}91928.3   & \cellcolor[HTML]{F9E3E1}52.6 \\
Hungarian                      & hun\_Latn & {\color[HTML]{60A5FA} High} & \cellcolor[HTML]{FFFEFE}0.14 & \cellcolor[HTML]{FFFEFE}646556  & \cellcolor[HTML]{FFFFFF}90517.8   & \cellcolor[HTML]{BBE4D0}85.8 \\
Sinhala (Latin)                & sin\_Latn & {\color[HTML]{FB7185} Low}  & \cellcolor[HTML]{FFFFFF}0.13 & \cellcolor[HTML]{FFFDFD}675122  & \cellcolor[HTML]{FFFFFF}87765.9   & \cellcolor[HTML]{F2BFBA}32.1 \\
West Central Oromo             & gaz\_Latn & {\color[HTML]{FB7185} Low}  & \cellcolor[HTML]{FFFEFE}0.14 & \cellcolor[HTML]{FFFFFF}615207  & \cellcolor[HTML]{FFFFFF}86129.0   & \cellcolor[HTML]{F1B8B3}28.1 \\
Haitian Creole                 & hat\_Latn & {\color[HTML]{FB7185} Low}  & \cellcolor[HTML]{FFFFFF}0.13 & \cellcolor[HTML]{FFFEFD}657970  & \cellcolor[HTML]{FFFFFF}85536.1   & \cellcolor[HTML]{FCF2F2}61.4 \\
Kazakh                         & kaz\_Cyrl & {\color[HTML]{60A5FA} High} & \cellcolor[HTML]{FFFBFB}0.16 & \cellcolor[HTML]{F6FBF9}534010  & \cellcolor[HTML]{FFFFFF}85441.6   & \cellcolor[HTML]{FFFFFF}68.3 \\
Urdu                           & urd\_Arab & {\color[HTML]{FB7185} Low}  & \cellcolor[HTML]{FFFBFB}0.16 & \cellcolor[HTML]{F5FBF8}527972  & \cellcolor[HTML]{FFFFFF}84475.5   & \cellcolor[HTML]{ECF7F2}73.4 \\
Thai                           & tha\_Thai & {\color[HTML]{60A5FA} High} & \cellcolor[HTML]{FFFEFE}0.14 & \cellcolor[HTML]{FFFFFF}603074  & \cellcolor[HTML]{FEFEFE}84430.4   & \cellcolor[HTML]{C8E9D9}82.6 \\
Luo                            & luo\_Latn & {\color[HTML]{FB7185} Low}  & \cellcolor[HTML]{EEF8F3}0.12 & \cellcolor[HTML]{FFFDFD}687087  & \cellcolor[HTML]{FEFEFE}82450.4   & \cellcolor[HTML]{F0B2AD}24.7 \\
Somali                         & som\_Latn & {\color[HTML]{FB7185} Low}  & \cellcolor[HTML]{FFFEFE}0.14 & \cellcolor[HTML]{FDFEFD}580127  & \cellcolor[HTML]{FEFEFE}81217.8   & \cellcolor[HTML]{F2BFBA}32.0 \\
Georgian                       & kat\_Geor & {\color[HTML]{FB7185} Low}  & \cellcolor[HTML]{FDF5F4}0.20 & \cellcolor[HTML]{E4F4EC}400695  & \cellcolor[HTML]{FDFEFE}80139.0   & \cellcolor[HTML]{F9E4E2}53.3 \\
Plateau Malagasy               & plt\_Latn & {\color[HTML]{FB7185} Low}  & \cellcolor[HTML]{FFFFFF}0.13 & \cellcolor[HTML]{FDFEFE}586134  & \cellcolor[HTML]{FDFEFD}76197.4   & \cellcolor[HTML]{F4C6C2}36.0 \\
Latvian                        & lvs\_Latn & {\color[HTML]{60A5FA} High} & \cellcolor[HTML]{FFFFFF}0.13 & \cellcolor[HTML]{FDFEFD}582382  & \cellcolor[HTML]{FCFEFD}75709.7   & \cellcolor[HTML]{C4E7D6}83.6 \\
Czech                          & ces\_Latn & {\color[HTML]{60A5FA} High} & \cellcolor[HTML]{FFFFFF}0.13 & \cellcolor[HTML]{F9FCFB}557529  & \cellcolor[HTML]{FCFDFC}72478.8   & \cellcolor[HTML]{AEDEC6}89.3 \\
Western Persian                & pes\_Arab & {\color[HTML]{60A5FA} High} & \cellcolor[HTML]{FFFEFE}0.14 & \cellcolor[HTML]{F3FAF6}506687  & \cellcolor[HTML]{FBFDFC}70936.2   & \cellcolor[HTML]{FEFDFD}67.4 \\
Ilocano                        & ilo\_Latn & {\color[HTML]{FB7185} Low}  & \cellcolor[HTML]{EEF8F3}0.12 & \cellcolor[HTML]{FDFEFD}580320  & \cellcolor[HTML]{FBFDFC}69638.4   & \cellcolor[HTML]{F6D0CD}41.7 \\
Icelandic                      & isl\_Latn & {\color[HTML]{60A5FA} High} & \cellcolor[HTML]{FFFFFF}0.13 & \cellcolor[HTML]{F4FAF7}514387  & \cellcolor[HTML]{FAFDFC}66870.3   & \cellcolor[HTML]{FFFFFF}68.3 \\
Moroccan Arabic                & ary\_Arab & {\color[HTML]{FB7185} Low}  & \cellcolor[HTML]{EEF8F3}0.12 & \cellcolor[HTML]{F8FCFA}550284  & \cellcolor[HTML]{FAFDFB}66034.1   & \cellcolor[HTML]{FDF8F8}64.6 \\
Northern Uzbek                 & uzn\_Latn & {\color[HTML]{60A5FA} High} & \cellcolor[HTML]{FFFFFF}0.13 & \cellcolor[HTML]{F1F9F5}495798  & \cellcolor[HTML]{FAFDFB}64453.7   & \cellcolor[HTML]{EAF7F1}73.7 \\
Mesopotamian Arabic            & acm\_Arab & {\color[HTML]{FB7185} Low}  & \cellcolor[HTML]{EEF8F3}0.12 & \cellcolor[HTML]{F6FBF9}532738  & \cellcolor[HTML]{FAFCFB}63928.6   & \cellcolor[HTML]{F0F9F5}72.3 \\
Slovak                         & slk\_Latn & {\color[HTML]{60A5FA} High} & \cellcolor[HTML]{EEF8F3}0.12 & \cellcolor[HTML]{F4FAF7}519748  & \cellcolor[HTML]{F9FCFB}62369.8   & \cellcolor[HTML]{D9F0E4}78.3 \\
Serbian                        & srp\_Cyrl & {\color[HTML]{FB7185} Low}  & \cellcolor[HTML]{FFFFFF}0.13 & \cellcolor[HTML]{EEF8F3}475857  & \cellcolor[HTML]{F9FCFB}61861.4   & \cellcolor[HTML]{B4E1CB}87.6 \\
Sundanese                      & sun\_Latn & {\color[HTML]{FB7185} Low}  & \cellcolor[HTML]{EEF8F3}0.12 & \cellcolor[HTML]{F3FAF6}510165  & \cellcolor[HTML]{F9FCFB}61219.8   & \cellcolor[HTML]{FAE7E5}54.8 \\
Tosk Albanian                  & als\_Latn & {\color[HTML]{60A5FA} High} & \cellcolor[HTML]{FFFFFF}0.13 & \cellcolor[HTML]{EDF7F2}462449  & \cellcolor[HTML]{F9FCFA}60118.4   & \cellcolor[HTML]{CDEBDC}81.2 \\
Lingala                        & lin\_Latn & {\color[HTML]{FB7185} Low}  & \cellcolor[HTML]{EEF8F3}0.12 & \cellcolor[HTML]{F1F9F5}495406  & \cellcolor[HTML]{F8FCFA}59448.7   & \cellcolor[HTML]{F1B7B3}27.9 \\
Kabuverdianu                   & kea\_Latn & {\color[HTML]{FB7185} Low}  & \cellcolor[HTML]{EEF8F3}0.12 & \cellcolor[HTML]{EDF7F2}463288  & \cellcolor[HTML]{F8FCFA}55594.6   & \cellcolor[HTML]{FBEEEC}58.7 \\
Cebuano                        & ceb\_Latn & {\color[HTML]{FB7185} Low}  & \cellcolor[HTML]{EEF8F3}0.12 & \cellcolor[HTML]{E9F6F0}437536  & \cellcolor[HTML]{F7FBF9}52504.3   & \cellcolor[HTML]{E7F5EE}74.7 \\
Estonian                       & est\_Latn & {\color[HTML]{60A5FA} High} & \cellcolor[HTML]{EEF8F3}0.12 & \cellcolor[HTML]{E8F5EF}430745  & \cellcolor[HTML]{F7FBF9}51689.4   & \cellcolor[HTML]{E1F3EA}76.2 \\
Korean                         & kor\_Hang & {\color[HTML]{60A5FA} High} & \cellcolor[HTML]{DFF2E8}0.11 & \cellcolor[HTML]{EDF8F3}468108  & \cellcolor[HTML]{F7FBF9}51491.9   & \cellcolor[HTML]{B3E1CB}87.8 \\
Afrikaans                      & afr\_Latn & {\color[HTML]{60A5FA} High} & \cellcolor[HTML]{EEF8F3}0.12 & \cellcolor[HTML]{E7F5EE}424235  & \cellcolor[HTML]{F6FBF9}50908.2   & \cellcolor[HTML]{B6E2CC}87.2 \\
Macedonian                     & mkd\_Cyrl & {\color[HTML]{60A5FA} High} & \cellcolor[HTML]{FFFFFF}0.13 & \cellcolor[HTML]{E3F3EB}390445  & \cellcolor[HTML]{F6FBF9}50757.9   & \cellcolor[HTML]{C2E7D5}84.0 \\
Guarani                        & grn\_Latn & {\color[HTML]{FB7185} Low}  & \cellcolor[HTML]{EEF8F3}0.12 & \cellcolor[HTML]{E7F5EE}418942  & \cellcolor[HTML]{F6FBF9}50273.0   & \cellcolor[HTML]{F3C2BE}34.0 \\
Finnish                        & fin\_Latn & {\color[HTML]{60A5FA} High} & \cellcolor[HTML]{EEF8F3}0.12 & \cellcolor[HTML]{E4F4EC}397741  & \cellcolor[HTML]{F6FBF8}47728.9   & \cellcolor[HTML]{D8F0E4}78.4 \\
Japanese                       & jpn\_Jpan & {\color[HTML]{60A5FA} High} & \cellcolor[HTML]{DFF2E8}0.11 & \cellcolor[HTML]{E8F6EF}432197  & \cellcolor[HTML]{F6FBF8}47541.7   & \cellcolor[HTML]{D9F0E5}78.2 \\
Danish                         & dan\_Latn & {\color[HTML]{60A5FA} High} & \cellcolor[HTML]{DFF2E8}0.11 & \cellcolor[HTML]{E8F5EF}427814  & \cellcolor[HTML]{F5FBF8}47059.5   & \cellcolor[HTML]{AFDFC7}89.0 \\
Romanian                       & ron\_Latn & {\color[HTML]{60A5FA} High} & \cellcolor[HTML]{EEF8F3}0.12 & \cellcolor[HTML]{E2F3EB}386978  & \cellcolor[HTML]{F5FBF8}46437.4   & \cellcolor[HTML]{ADDEC6}89.4 \\
Ukrainian                      & ukr\_Cyrl & {\color[HTML]{60A5FA} High} & \cellcolor[HTML]{FFFFFF}0.13 & \cellcolor[HTML]{DEF1E8}356123  & \cellcolor[HTML]{F5FBF8}46296.0   & \cellcolor[HTML]{B5E2CC}87.3 \\
Lithuanian                     & lit\_Latn & {\color[HTML]{60A5FA} High} & \cellcolor[HTML]{EEF8F3}0.12 & \cellcolor[HTML]{E2F3EB}384397  & \cellcolor[HTML]{F5FBF8}46127.6   & \cellcolor[HTML]{C3E7D5}83.9 \\
Hebrew                         & heb\_Hebr & {\color[HTML]{60A5FA} High} & \cellcolor[HTML]{DFF2E8}0.11 & \cellcolor[HTML]{E6F5ED}414788  & \cellcolor[HTML]{F5FBF8}45626.7   & \cellcolor[HTML]{C1E6D4}84.4 \\
Javanese                       & jav\_Latn & {\color[HTML]{FB7185} Low}  & \cellcolor[HTML]{DFF2E8}0.11 & \cellcolor[HTML]{E5F4ED}406097  & \cellcolor[HTML]{F5FBF8}44670.7   & \cellcolor[HTML]{E1F3EA}76.1 \\
Dutch                          & nld\_Latn & {\color[HTML]{60A5FA} High} & \cellcolor[HTML]{DFF2E8}0.11 & \cellcolor[HTML]{E4F4EC}397947  & \cellcolor[HTML]{F5FBF8}43774.2   & \cellcolor[HTML]{B2E0CA}88.1 \\
Russian                        & rus\_Cyrl & {\color[HTML]{60A5FA} High} & \cellcolor[HTML]{EEF8F3}0.12 & \cellcolor[HTML]{E0F2E9}370275  & \cellcolor[HTML]{F5FBF8}43559.8   & \cellcolor[HTML]{A7DCC2}90.9 \\
Bulgarian                      & bul\_Cyrl & {\color[HTML]{60A5FA} High} & \cellcolor[HTML]{EEF8F3}0.12 & \cellcolor[HTML]{DFF2E8}362387  & \cellcolor[HTML]{F5FAF8}43486.4   & \cellcolor[HTML]{ABDDC5}89.9 \\
Croatian                       & hrv\_Latn & {\color[HTML]{60A5FA} High} & \cellcolor[HTML]{EEF8F3}0.12 & \cellcolor[HTML]{DFF2E8}360991  & \cellcolor[HTML]{F5FAF8}43318.9   & \cellcolor[HTML]{C1E6D4}84.3 \\
Slovenian                      & slv\_Latn & {\color[HTML]{60A5FA} High} & \cellcolor[HTML]{EEF8F3}0.12 & \cellcolor[HTML]{DFF2E8}360206  & \cellcolor[HTML]{F5FAF8}43224.7   & \cellcolor[HTML]{B6E2CC}87.2 \\
Traditional Chinese            & zho\_Hant & {\color[HTML]{60A5FA} High} & \cellcolor[HTML]{DFF2E8}0.11 & \cellcolor[HTML]{E3F3EB}392790  & \cellcolor[HTML]{F5FAF8}43206.9   & \cellcolor[HTML]{B0DFC8}88.8 \\
Italian                        & ita\_Latn & {\color[HTML]{60A5FA} High} & \cellcolor[HTML]{DFF2E8}0.11 & \cellcolor[HTML]{DFF2E9}366630  & \cellcolor[HTML]{F4FAF7}40329.3   & \cellcolor[HTML]{AADDC4}90.2 \\
Polish                         & pol\_Latn & {\color[HTML]{60A5FA} High} & \cellcolor[HTML]{EEF8F3}0.12 & \cellcolor[HTML]{DBF0E6}333593  & \cellcolor[HTML]{F4FAF7}40031.2   & \cellcolor[HTML]{C2E7D5}84.0 \\
Vietnamese                     & vie\_Latn & {\color[HTML]{60A5FA} High} & \cellcolor[HTML]{DFF2E8}0.11 & \cellcolor[HTML]{DEF2E8}359211  & \cellcolor[HTML]{F4FAF7}39513.2   & \cellcolor[HTML]{B0DFC8}88.6 \\
Norwegian Bokmal               & nob\_Latn & {\color[HTML]{FB7185} Low}  & \cellcolor[HTML]{DFF2E8}0.11 & \cellcolor[HTML]{DEF1E8}355727  & \cellcolor[HTML]{F4FAF7}39130.0   & \cellcolor[HTML]{ABDDC4}90.1 \\
Turkish                        & tur\_Latn & {\color[HTML]{60A5FA} High} & \cellcolor[HTML]{DFF2E8}0.11 & \cellcolor[HTML]{DDF1E7}349338  & \cellcolor[HTML]{F3FAF7}38427.2   & \cellcolor[HTML]{BDE5D1}85.4 \\
Simplified Chinese             & zho\_Hans & {\color[HTML]{60A5FA} High} & \cellcolor[HTML]{CFEBDD}0.10 & \cellcolor[HTML]{DFF2E8}362454  & \cellcolor[HTML]{F3FAF6}36982.7   & \cellcolor[HTML]{ACDEC6}89.6 \\
Waray                          & war\_Latn & {\color[HTML]{FB7185} Low}  & \cellcolor[HTML]{EEF8F3}0.12 & \cellcolor[HTML]{D7EFE3}306588  & \cellcolor[HTML]{F3FAF6}36790.6   & \cellcolor[HTML]{FBEFEE}59.3 \\
Swedish                        & swe\_Latn & {\color[HTML]{60A5FA} High} & \cellcolor[HTML]{DFF2E8}0.11 & \cellcolor[HTML]{DAF0E5}330028  & \cellcolor[HTML]{F3FAF6}36303.1   & \cellcolor[HTML]{ACDEC6}89.6 \\
Spanish                        & spa\_Latn & {\color[HTML]{60A5FA} High} & \cellcolor[HTML]{DFF2E8}0.11 & \cellcolor[HTML]{DAF0E5}329321  & \cellcolor[HTML]{F3FAF6}36225.3   & \cellcolor[HTML]{A8DCC2}90.8 \\
North Levantine Arabic         & apc\_Arab & {\color[HTML]{FB7185} Low}  & \cellcolor[HTML]{DFF2E8}0.11 & \cellcolor[HTML]{D8EFE4}315102  & \cellcolor[HTML]{F2FAF6}34661.2   & \cellcolor[HTML]{E5F5ED}75.1 \\
Catalan                        & cat\_Latn & {\color[HTML]{60A5FA} High} & \cellcolor[HTML]{DFF2E8}0.11 & \cellcolor[HTML]{D7EFE3}306329  & \cellcolor[HTML]{F2FAF6}33696.2   & \cellcolor[HTML]{ACDEC5}89.7 \\
German                         & deu\_Latn & {\color[HTML]{60A5FA} High} & \cellcolor[HTML]{DFF2E8}0.11 & \cellcolor[HTML]{D6EEE2}297520  & \cellcolor[HTML]{F2F9F6}32727.2   & \cellcolor[HTML]{A3DABF}92.1 \\
Najdi Arabic                   & ars\_Arab & {\color[HTML]{FB7185} Low}  & \cellcolor[HTML]{EEF8F3}0.12 & \cellcolor[HTML]{D0ECDE}253562  & \cellcolor[HTML]{F1F9F5}30427.4   & \cellcolor[HTML]{E5F5ED}75.2 \\
Basque                         & eus\_Latn & {\color[HTML]{60A5FA} High} & \cellcolor[HTML]{EEF8F3}0.12 & \cellcolor[HTML]{D0ECDE}253348  & \cellcolor[HTML]{F1F9F5}30401.8   & \cellcolor[HTML]{FEFDFD}67.6 \\
Tagalog                        & tgl\_Latn & {\color[HTML]{60A5FA} High} & \cellcolor[HTML]{FFFFFF}0.13 & \cellcolor[HTML]{CDEADC}227736  & \cellcolor[HTML]{F1F9F5}29605.7   & \cellcolor[HTML]{E7F6EE}74.6 \\
French                         & fra\_Latn & {\color[HTML]{60A5FA} High} & \cellcolor[HTML]{DFF2E8}0.11 & \cellcolor[HTML]{D1ECDF}259581  & \cellcolor[HTML]{F1F9F5}29420.4   & \cellcolor[HTML]{A4DBC0}91.7 \\
Egyptian Arabic                & arz\_Arab & {\color[HTML]{FB7185} Low}  & \cellcolor[HTML]{EEF8F3}0.12 & \cellcolor[HTML]{CCEADB}224620  & \cellcolor[HTML]{F1F9F5}26954.4   & \cellcolor[HTML]{CDEBDC}81.3 \\
Indonesian                     & ind\_Latn & {\color[HTML]{60A5FA} High} & \cellcolor[HTML]{EEF8F3}0.12 & \cellcolor[HTML]{CCEADB}223840  & \cellcolor[HTML]{F1F9F5}26860.8   & \cellcolor[HTML]{BEE5D2}85.0 \\
Modern Standard Arabic         & arb\_Arab & {\color[HTML]{60A5FA} High} & \cellcolor[HTML]{EEF8F3}0.12 & \cellcolor[HTML]{CBEADB}217531  & \cellcolor[HTML]{F0F9F5}26103.7   & \cellcolor[HTML]{CFECDE}80.8 \\
Standard Malay                 & zsm\_Latn & {\color[HTML]{60A5FA} High} & \cellcolor[HTML]{EEF8F3}0.12 & \cellcolor[HTML]{CBEADB}217255  & \cellcolor[HTML]{F0F9F5}26070.6   & \cellcolor[HTML]{B2E0CA}88.1 \\
Portuguese                     & por\_Latn & {\color[HTML]{60A5FA} High} & \cellcolor[HTML]{DFF2E8}0.11 & \cellcolor[HTML]{CCEADB}223499  & \cellcolor[HTML]{F0F9F4}24584.9   & \cellcolor[HTML]{AADDC4}90.2 \\
English                        & eng\_Latn & {\color[HTML]{60A5FA} High} & \cellcolor[HTML]{CFEBDD}0.10 & \cellcolor[HTML]{C4E7D6}167621  & \cellcolor[HTML]{EEF8F3}17588.0   & \cellcolor[HTML]{99D6B8}94.6\\
\bottomrule
\caption{Full results for all 122 languages by Qwen3-8B on Belebele.
J/tok reports the energy per output token.
Output Tokens reports the total number of tokens generated across 900 responses.
}
\label{tab:full_results}
\end{longtable}
\endgroup
\twocolumn

\section{Language List and Energy Statistics}
\label{sec:appendix_languages}

Table~\ref{tab:full_results} provides the complete list of all 122 Belebele languages with their resource level, energy per output token, total output tokens, total energy consumption, and task accuracy on Qwen3-8B under zero-shot CoT prompting.

\subsection{Discussion}
\label{sec:table_discussion}

Table~\ref{tab:full_results} surfaces several patterns that we briefly highlight here.

Apart from the analysis we conduct in \S~\ref{subsec: energy-vary-across-languages} that there is significant energy gap across languages, the languages with the highest output-token counts also exhibit the highest J/tok, and energy cost and accuracy are inversely related, we also note several other observations here.

Several languages labeled high-resource (Zulu $28.7$, Tswana $29.2$, Xhosa $29.8$, Bengali-Latin $29.9$) underperform languages labeled low-resource (Marathi $76.4$, Javanese $76.1$, Kabuverdianu $58.7$).
Therefore, \textit{the High/Low resource binary is a coarse predictor.}
Script familiarity and overlap with the pretraining mixture are plausible explanations for the variance in accuracy on this benchmark.

Latin transliterations reduce token counts substantially, but their effect on accuracy is inconsistent.
Hindi-Latin loses only modestly relative to Devanagari ($75.9 \to 66.7$), whereas Bengali-Latin collapses from $78.3$ to $29.9$ and Sinhala-Latin from $51.8$ to $32.1$.
Romanization helps the tokenizer but can disrupt the model's lexical and morphological knowledge of the language.
Therefore, \textit{romanization is not a uniform win.}

There are outliers to our analysis.
Georgian (\texttt{kat\_Geor}) exhibits a relatively high J/tok ($0.20$) at a moderate output length ($\sim\!4.0\times 10^{5}$ tokens) and lands at $53.3\%$ accuracy, which is an unusual position relative to other non-Latin scripts.
This likely reflects intrinsic properties of the language, including its agglutinative morphology with up to eight verbal morphemes and polypersonal agreement~\citep{aronson1990georgian, hewitt1995georgian}, and the relative isolation of the Kartvelian family and its unique Mkhedruli script, which shares no genetic relationship with other major writing systems~\citep{boeder2005south}.

\Cref{fig:world_maps_all} plots the geographic distribution of the results from \Cref{tab:full_results}.
Note that the country-fill assignment necessarily simplifies multilingual states (e.g.\ Belgium is shown by Dutch, Switzerland by German, Mali by Bambara rather than the colonial language), and pin positions are illustrative rather than cartographically precise.
However, we can still observe that Europe, the Americas, East Asia, and the anglophone parts of Africa and Oceania appear uniformly dark-green and pale-red (high accuracy while low energy cost), while the band running from West Africa through the Sahel, the Horn of Africa, the Himalayan plateau, and mainland Southeast Asia appears pale-green and dark-red (low accuracy while high energy cost).

\section{Experimental Setups}
\label{appsec: experimental-setups}

\paragraph{Belebele zero-shot CoT prompt construction.}
For Belebele \citep{bandarkar-etal-2024-belebele}, we use a zero-shot chain-of-thought prompt covering all 122 languages of the Belebele test split.
The prompt format, identical across languages up to translation, is:
\begin{center}
\small
\fbox{\parbox{0.93\columnwidth}{\textit{<per-language instruction, ending with the final line ``\#\#\#\# N''>}\\[2pt]
\textit{<P-label>: <passage>}\\
\textit{<Q-label>: <question>}\\
\textit{<O-label>: 1. <option 1> 2. <option 2> 3. <option 3> 4. <option 4>}\\
\textit{<per-language primer ``Let's think step by step.''>}}}
\end{center}
The per-language instruction is a translation of the canonical English instruction
\emph{``Read the passage and answer the multiple-choice question. Reason step by step. Then, on a new last line, write the number of the correct option (1, 2, 3, or 4) in exactly this format:''}
followed by \texttt{\#\#\#\# N} on its own final line, which is used in every language as the parser anchor.
The primer is a translation of \emph{``Let's think step by step.''}.
The passage, question, and four answer options are taken directly from the Belebele test split for the target language.
In addition, the Passage/Question/Options labels are all translated to the target language.

\paragraph{Per-language translation pipeline.}
We translate the instruction and primer for each Belebele language using NLLB-200~\citep{costa2022no}, then back-translate \citep{rapp-2009-backtranslation, sennrich-etal-2016-improving, moon-etal-2020-revisiting} every candidate and filter by BERTScore \citep{Zhang2020BERTScore} $\geq 0.85$ and COMET \citep{rei-etal-2020-comet} $\geq 0.75$.
This automated pipeline yields translations for 93 of the 122 Belebele languages.
Of the remaining 29 languages, 1 is the source language (English) and requires no translation.
For the remaining 28 languages, we hand-curate the instruction and primer using Claude-4.7-Opus\footnote{\url{https://www.anthropic.com/news/claude-opus-4-7}} and manually verify each back-translation.
These 28 cases comprise 11 languages with dropped clauses (e.g.\ Bulgarian, French, Russian, Ukrainian), 10 with mistranslations (e.g.\ Korean, Bambara, Igbo), 1 partial (Fulfulde), and 6 romanized Latin-script Belebele variants for which NLLB has no direct code (\texttt{arb\_Latn}, \texttt{ben\_Latn}, \texttt{hin\_Latn}, \texttt{npi\_Latn}, \texttt{sin\_Latn}, \texttt{urd\_Latn}).
In all manual cases, we translate only the prose and keep the parser anchor \texttt{\#\#\#\# N} as the final line.

\paragraph{Translated GSM8K and LM-Arena.}
Similarly, the translated GSM8K and LM-Arena prompts in \S~\ref{sec:lmarena} are produced with an NLLB-200 translation and back-translation quality-control pipeline (BERTScore and COMET), yielding 272 and 266 prompts that pass the quality bar in all 8 languages.
We translate the original GSM8K \citep{cobbe2021gsm8k} rather than directly use MGSM \citep{shi2023language} so the math task spans the identical 8-language subset.
All three tasks in \Cref{tab:cross_task} share the zero-shot prompt design: GSM8K appends the per-language ``Let's think step by step.'' primer to the translated question (similar to the zero-shot CoT for Belebele), while LM-Arena uses the translated user message directly (the standard open-chat).
To match the request count of the 900-request Belebele runs we issue each prompt three times, giving $272\times3=816$ and $266\times3=798$ effective requests, respectively.
Because per-token energy is invariant to the total request count by construction, the differing effective request counts across tasks (900, 816, and 798) do not affect comparability.

\paragraph{Serving configuration and parallelism.}
All models are served via vLLM~\citep{kwon2023efficient}, running inside a Singularity container, with the batch size (the maximum number of concurrent sequences) set to 256 unless otherwise noted.
For the cross-model experiment (\S~\ref{sec:cross_model}) on L40S GPUs, we run Qwen3-8B, Qwen3-14B, and Llama-3.1-8B-Instruct on a single GPU, Qwen3-32B and gemma-3-27b-it by pipeline parallelism with degree~2.
For a batch size of 512, we use a 1800-request run (two passes over the 900-prompt set).
Specifically, the 900-request run's steady-state window collapses because $900 < 2\times512$, whereas $1800 > 2\times512$ yields a valid window.
Per-token energy is request-count-invariant by construction in the saturated-serving regime \citep{chung2026the}, so this does not affect comparability across batch sizes.
\end{document}